\documentclass{article}

\usepackage{longtable,booktabs,array,makecell}
\usepackage[table]{xcolor} 
\usepackage{graphicx}%
\usepackage{multirow}%
\usepackage{caption}
\usepackage{amsmath,amssymb,amsfonts}%
\usepackage{amsthm}%
\usepackage{comment}
\usepackage{mathrsfs}%
\usepackage[title]{appendix}%
\usepackage{xcolor}%
\usepackage{textcomp}%
\usepackage{manyfoot}%
\usepackage{algorithm}%
\usepackage{algorithmicx}%
\usepackage{algpseudocode}%
\usepackage{listings}%
\usepackage{float}
\usepackage{tabularx}
\usepackage[skip=0pt]{caption}
\usepackage[a4paper, left=1in, right=1in, top=1in, bottom=1in]{geometry}
\usepackage[style=apa, backend=biber]{biblatex}%
\usepackage[colorlinks=true, linkcolor=blue, citecolor=blue]{hyperref}
\usepackage{ltablex}     
\usepackage{makecell}    
\usepackage{pifont}      
\usepackage{array}       

\newcommand{\cmark}{\textnormal{\ding{51}}}
\newcommand{\xmark}{\textnormal{\ding{53}}}

\newcolumntype{L}[1]{>{\raggedright\arraybackslash}p{#1}}
\title{Sample Title}

\author{
Md. Tawfique Ihsan* \\
Md. Rakibul Hasan Rafi* \\
Ahmed Shoyeb Raihan \\
Imtiaz Ahmed \\
Abdullahil Azeem
}
\date{} 

\title{Boosting Predictive Performance on Tabular Data through Data Augmentation with Latent-Space Flow-Based Diffusion}

\begin{document}

\newtheorem{theorem}{Theorem}
\newtheorem{proposition}[theorem]{Proposition}%
\newtheorem{example}{Example}%
\newtheorem{remark}{Remark}%

\newtheorem{definition}{Definition}%

\raggedbottom




\maketitle
\begingroup
\renewcommand\thefootnote{*}
\footnotetext{Md. Tawfique Ihsan and Md. Rakibul Hasan Rafi contributed equally to this work.}
\endgroup
\begin{abstract}
Severe class imbalance is pervasive in real-world tabular learning, where rare but critical minority classes (e.g., defects, failures, adverse events) are precisely those most needed for reliable prediction. While recent generative approaches—GANs, VAEs, and neural diffusion—can synthesize minority samples to rebalance training data and can improve minority-class prediction in class-imbalanced problems, they often struggle with tabular heterogeneity, training stability, and privacy risks. We introduce a family of latent-space, tree-driven diffusion methods for minority oversampling that couple conditional flow matching (CFM) with gradient-boosted trees (GBTs) as the vector-field learner and operate in compact latent spaces to preserve tabular structure and reduce computational cost. Concretely, we propose three novel approaches: \emph{PCAForest} (linear PCA embedding), \emph{EmbedForest} (learned nonlinear embedding), and \emph{AttentionForest} (attention-augmented embedding); each paired with a GBT-parameterized flow and a decoder back to the original feature space. The design aligns the generator with inductive biases that dominate tabular prediction while enabling training and sampling in lower-dimensional manifolds. We evaluate downstream utility (minority recall, precision, calibration), statistical similarity (Wasserstein distance), and empirical privacy via the nearest-neighbor distance ratio (NNDR) and distance-to-closest-record (DCR), and analyze robustness with ablations over embedding dimension and optimization settings. Aggregated across 11 datasets from healthcare, finance, and manufacturing, \textbf{AttentionForest} attains the best average minority-class performance while maintaining competitive precision/calibration and lower or comparable distributional divergence from real data; \textbf{PCAForest} and \textbf{EmbedForest} deliver near-parity utility with substantially lower generation time,upto 50\% offering favorable accuracy–efficiency trade-offs. Privacy by NNDR/DCR remains on par with or better than ForestDiffusion across the aggregate, indicating that latent-space training does not introduce measurable leakage under these probes. An ablation pooled over datasets shows that smaller embeddings generally boost minority recall, whereas aggressive learning rates harm stability; the methods remain robust across practical augmentation ranges. Taken together, latent-space, tree-driven diffusion offers a data-efficient and privacy-aware path to high-fidelity tabular augmentation under severe class imbalance.
\end{abstract}

\section{{Introduction}}\label{sec1}

A fundamental challenge of binary classification problems and anomaly detection is to deal with class imbalance. The class imbalance in binary classification refers to the situation in which one class of data heavily dominates the other in a dataset \parencite{ghosh2024class, seliya2021literature}. For example, the number of non-defective items in a manufacturing plant is higher than the defective items. The number of affected patients in the case of a rare disease is generally lower than the number of non-affected people. The number of fraudulent transactions is rare compared to the legitimate transactions in a transaction credibility report. In all of the cases, the number of instances where a particular attribute (defect, disease, fraudulence) is present is noticeably lower than the instances where it does not. These two instances are called positive class and negative class respectively.

The risk of performing classification tasks with this type of dataset is extremely high as the scarcity of negative class data may result in the class being undiscovered or ignored by the classifier. This inclination of the classifier towards the more frequent or positive class is known as imbalanced or rare class learning problem which can contribute to the suboptimal performance of the classifier \parencite{machado2022benchmarking}. 

To mitigate this problem, researchers have developed various methods that can be broadly classified into two groups: algorithm level methods and data level methods \parencite{sauber2022use, ma2025class}. The goal of algorithm level methods and data level methods is to redesign the learning algorithms and to reduce the imbalance in the dataset respectively in order to attenuate the bias of the classifier towards the majority class \parencite{sauber2022use}. \textcite{sauglam2022novel, abedin2023combining, dixit2023sampling} proposed hybrid methods for imbalanced learning by combining the advantages of algorithm level and data level approaches. However, this integration is associated with increased model complexity and extensive parameter training \parencite{chen2024survey}. Moreover, the use of hybrid approaches in diverse data modalities is underexplored \parencite{tsai2024hybrid}.

In terms of algorithm, one of the proposed methods is learning from one class or recognition-based learning instead of learning from two classes or discrimination-based learning \parencite{japkowicz1995novelty}. The objective of one class learning is to determine a decision boundary based on the positive class only and without explicitly visiting the negative class so that it recognizes as many positive classes as possible minimizing the recognition of negative classes \parencite{perera2021one}. The limitation of this method is that constructing a decision boundary based on only one class (positive class) is extremely difficult and can often lead to inaccurate boundary and increases false positives. Additionally, one class learning might be an impediment to feature selection as this method deals with single class only instead of two \parencite{seliya2021literature}. An alternative to one class classification is binary classification which requires the presence of two classes. Although binary classification poses some challenges such as being biased towards the majority class \parencite{bellinger2012one}, it shows promising result in Credit Card Fraud detection \parencite{leevy2023comparative}. Additionally, the modality is found to be the most important factor in selecting classification strategy \parencite{bellinger2017sampling}.

The second and most popular solution method for imbalanced learning problem from the perspective of imbalanced dataset is to reduce the 
imbalance in the dataset. The reduction can be done in two ways; either by removing majority class data or by adding data to the minority class. The data removal procedure is known as random under-sampling in which the majority class is down-sized to match the size of the minority class \parencite{chaipanha2022smote}. Despite being useful for a large dataset, random under-sampling method might eliminate some important data necessary for the classifier \parencite{mohammed2020machine, wongvorachan2023comparison}. This limitation can be abated with the addition of data to the minority class to reduce the imbalance. One of the data integration methods gradually increases the size of the minority class through a process called re-sampling which is also known as over-sampling \parencite{tarawneh2022stop} and can be done at random or directed \parencite{mohammed2020machine}. As the over-sampling method adds duplicated data to the minority class without creating new instances, the chance of over-fitting increases \parencite{mease2007boosted, ganganwar2012overview, wongvorachan2023comparison} along with time consuming learning procedure if the dataset is large \parencite{drummond2003c4, santos2018cross}. However, when compared with over-sampling, random under-sampling performs better if the dataset is moderately imbalanced \parencite{mohammed2020machine}. The selection of sampling technique largely depends on dataset properties \parencite{shamsudin2020combining,loyola2016study}.

The conundrum of selecting appropriate and effective sampling technique for diverse datasets motivated researchers to reduce the imbalance in the dataset by adding synthetic data to the minority class. Synthetic data is defined as the artificially generated data by an Artificial Intelligence (AI) algorithm that has been trained on real data and exhibits the characteristics of real data \parencite{lucini2022real}. In quest of data generation \textcite{chawla2002smote} developed SMOTE (Synthetic Minority Over-sampling Technique) which is a method of selecting minority class instance and its closest neighbors to generate new samples through interpolation. Since then SMOTE has become a pioneering algorithm for synthetic data generation in imbalanced learning \parencite{fernandez2018smote}. SMOTE generalizes the decision boundary and does not allow significant reduction in performance of the majority class \parencite{han2005borderline}. However, SMOTE might induce overgeneralization as it produces new samples with positive class (minority class) label and overlooks the negative class (majority class) label \parencite{branco2016survey}.  

Another significant progress in synthetic data generation was achieved when 
\textcite{goodfellow2014generative} proposed Generative Adversarial Networks (GAN) which generates new samples by simultaneous training of two multilayer perceptrons; Generator and Discriminator. Analogous to the game theory, generator and discriminator always competes with each other as two player minimax game until the samples generated by the generator can no longer be distinguished from the original sample by discriminator. The data generated by this iterative update of generator and discriminator networks exhibits excellent resemblance of features in comparison with the original data. Despite being a major step forward in imbalanced learning to produce high quality samples in various types of data, GANs are subject to some challenges such as vanishing gradient problem \parencite{zhang2019towards}, mode collapse \parencite{dhariwal2021diffusion} and non-convergence \parencite{mescheder2018training, sharma2024generative}.

Faced with these limitations of GANs, researchers developed likelihood-based models to produce GAN-like samples \parencite{child2020very, de2019hierarchical, nash2021generating}. The advantage of these models is that these models are capable of capturing more diversity. In addition to that, training and scaling these models are easier than GANs. However, these likelihood-based models does not provide accurate visual representation in terms of image data and suffers from slow sampling rate compared to GANs \parencite{dhariwal2021diffusion}. Addressing the limitations another model called Denoising Diffusion Probabilistic Model (DDPM) was presented by \textcite{ho2020denoising} and later modified by \textcite{nichol2021improved} that can generate high-quality samples and offers easy scalability, better mode coverage and comparatively fast sampling as well. Moreover, the training objective remains stationary and well-grounded in DDPM \parencite{dhariwal2021diffusion}. DDPM is a refinement of the original diffusion probabilistic model or diffusion model in short. Proposed by \parencite{sohl2015deep}, diffusion model showed the ability to generate high quality samples often outperforming other generative models like GANs . In diffusion model, gaussian noise are gradually added to a data distribution until it resembles a pure noise. Then the original data distribution is learned through a reverse process called denoising and samples similar to the original data are produced. Diffusion models offer good log-likelihoods, excellent quality samples and comparatively fast sampling. 

Since the development of diffusion model, most of the work with this generative approach has been used with image data in various domains such as colorization \parencite{song2020score}, segmentation \parencite{baranchuk2021label}, super resolution \parencite{saharia2022photorealistic}. Apart from image data diffusion models have also been researched in other fields including Natural Language Processing (NLP), time-series and molecular graphs. But the application of tabular data in synthetic data generation specifically with diffusion model is least explored \parencite{kadra2021well, fakoor2020fast}.

Tabular data is defined as a subset of heterogeneous data that is generally organized in a table. Each row of the table represents a unique data point and each column is portrayed as different features \parencite{borisov2022deep}. Tabular data, which is also known as structured data \parencite{ryan2020deep}, is one of the ancient forms of data. Before the widespread digital collection of image, text and sound, almost all of data available were in tabular format \parencite{jdanov2019human}. Consequently, tabular data was initially used at the outset of machine learning research \parencite{borisov2022deep}. It is still the most common type of data in real world AI \parencite{bughin2018notes}. Despite being very popular and ubiquitous, tabular data possess some fundamental challenges unlike image and text data. \textcite{kim2022stasy} argued that tabular data has some exceptional properties that makes it extremely difficult to model probabilistically. Tabular data is strained with the presence of missing values \parencite{sanchez2020improving} and outliers \parencite{pang2021deep}. In addition to this, inductive bias makes it more difficult to work with tabular data owing to the fact that, Convolutional Neural Networks (CNN) and Recurrent Neural Networks (RNN) assume that spatial and temporal relationships exist between data points. Since tabular data lacks such relationship, working with it becomes an arduous task \parencite{borisov2023deeptlf}. Moreover, the presence of categorical features makes tabular data even more challenging to work with \parencite{hancock2020survey} without pre-processing which might cause information loss \parencite{borisov2022deep}. The intricacies of tabular data makes it formidable to use it in generative models, particularly in diffusion model.

\textcite{kotelnikov2023tabddpm} showed that although tabular data possess these complexities, diffusion model can still learn the distribution of tabular data to yield superior performance in terms of data generation. However, the limitation of diffusion model is that it entails complete or fully observed data for training. \textcite{jolicoeur2024generating} proposed Forest-Diffusion which is a diffusion based data generation and imputation method that does not only excel at handling missing data but also produce high quality synthetic data. Forest-diffusion uses XGBoost, a popular Gradient Boosted Tree (GBT) method, as universal function estimator instead of neural networks. GBTs are a collection of sequential decision trees where error of the previous tree is used to train the later. GBTs perform better in tabular data whereas neural networks are more suitable for unstructured data like image, audio, text etc. for prediction and classification \parencite{shwartz2022tabular}. For this advantage, GBTs have been used in forest-diffusion for synthetic data generation. Even though diffusion models offer multiple benefits, one of the most prominent shortcomings it has is the positive correlation between data generation efficiency and training time. The training time plays a crucial role in achieving a certain level of efficiency in data generation \parencite{franzese2023much, yang2023diffusion}. Moreover, handling large dataset with diffusion model often leads to prolonged training time which in turn can cause excessive memory usage and be a deterrent to achieve the expected states \parencite{cao2024survey}.

For most of the generative models including diffusion model and GANs, longer training time and high computational demand have always been a hindrance to achieve optimal performance that is to generate high quality samples. In contrast, generating high quality samples often requires rigorous computation with sophisticated hardware. Therefore, multiple studies have been conducted parallel to generative models on how to obtain optimum results at low computational cost and increase computational efficiency \parencite{asperti2023comparing,asru2025automation}. One strategy is to reduce the dimension of the data by projecting it into a latent space in order to decrease computational cost \parencite{mittal2021symbolic}. Generative models can perform efficiently if low-dimensional latent space is properly utilized \parencite{vahdat2021score}. Latent space transformation makes the training and sampling of generative models computationally efficient by capturing the underlying structure and features of data in a compressed form. It also enables the researchers to include large and diverse datasets. The latent space representation for diffusion model has been used in various domains such as image synthesis \parencite{rombach2022high}, text generation \parencite{lovelace2023latent}, text-to-audio generation \parencite{liu2023audioldm}. Nevertheless, there is a dearth of studies on latent space representation for gradient boosted tree based diffusion model in synthetic data generation. Moreover, tabular data generation in latent space has also not been studied extensively.

In this work, we introduce a latent-space, tree-driven diffusion framework for synthetic minority oversampling in tabular data. The core idea is to learn the diffusion vector field with gradient-boosted trees under conditional flow matching while performing both training and sampling in a compact latent space defined by linear or learned encoders. We instantiate three concrete variants tailored to different fidelity--efficiency trade-offs: PCAForest, which uses a PCA encoder/decoder; EmbedForest, which employs a lightweight autoencoder to learn a nonlinear latent manifold; and AttentionForest, which couples a transformer autoencoder with Forest-Diffusion to model higher-order feature interactions. Operating in latent space aims to preserve tabular structure, align the generator with inductive biases common in tabular prediction, and reduce computational cost. We place these methods within a unified, reproducible protocol that fixes a real-only test split and evaluates downstream utility (minority recall, precision, calibration), statistical similarity (Wasserstein distance), and empirical privacy (nearest-neighbor distance ratio, NNDR; distance-to-closest-record, DCR) over specific augmentation ratios. Our contributions are summarized as follows.

\begin{itemize}
    \item We formulate a latent-space Forest-Diffusion framework that learns the diffusion vector field with GBTs under conditional flow matching and supports end-to-end training and sampling via encoder--decoder mappings.
    \item We propose three variants---\emph{PCAForest}, \emph{EmbedForest}, and \emph{AttentionForest}---realizing linear (PCA) and learned (autoencoder / transformer autoencoder) latent geometries to span accuracy--efficiency trade-offs for tabular synthesis.
    \item We design an end-to-end pipeline that (i) isolates minority-class samples, (ii) encodes them into a compact latent space, (iii) trains a GBT-driven flow, and (iv) decodes to produce synthetic minority records; the same latent representations can optionally support downstream classification in latent space for additional efficiency.
    \item We establish a rigorous evaluation protocol that fixes a real-only 30\% test split, sweeps augmentation ratios from 25\% to 300\%, and benchmarks against SMOTE, CTGAN, and Forest-Diffusion using (a) downstream utility with \emph{Random Forest} and \emph{XGBoost} classifiers (b) distributional similarity, and (c) empirical privacy; we apply this protocol to a diverse suite of tabular datasets spanning multiple domains and feature scales.
    \item We conduct diagnostic analyses, including ablations over embedding dimension, optimization settings (e.g., learning rate), and architectural choices (PCA vs.\ autoencoder vs.\ transformer autoencoder), and we profile computational cost to characterize fidelity--efficiency trade-offs.
\end{itemize}

\section{{Related work}}\label{sec2}

Research on imbalanced learning problem intensified since the beginning of the current century when two workshops at the AAAI and ICML conferences were held in 2000 \parencite{japkowicz2000learning} and 2003 \parencite{japkowicz2003class} respectively. Before that, there were no integrated research work in this domain. \textcite{japkowicz2000learning} also presented a comprehensive comparison of different methods proposed to attenuate the imbalanced learning problem. The introduction of SMOTE marked a significant early breakthrough in imbalanced learning, utilizing oversampling to address class imbalance \parencite{chawla2002smote}. SMOTE is considered as the foundation of oversampling with artificial data generation. Numerous methodologies have been proposed building upon the SMOTE algorithm to address its challenges and enhance its performance. Selecting suitable minority class instances is a usual practice in SMOTE. \textcite{han2005borderline} proposed Borderline-SMOTE which uses border points (minority class instances surrounded by both majority and minority class) for oversampling and neglects noise points (minority class instances surrounded by majority class mostly), thus avoiding the outliers. Safe-Level-SMOTE, another extension of SMOTE, allocates safe level to each of its minority class samples before generating synthetic data and synthetic data are generated only in the region closer to the highest safe level \parencite{bunkhumpornpat2009safe}. \textcite{fernandez2018smote} summarized some other variants of SMOTE algorithms based on different interpolation methods, dimension reduction, relabeling, noisy sample filtration. Adaptive synthetic sampling approach (ADASYN) uses difficult to learn minority class samples to generate synthetic data \parencite{he2008adasyn}. Although SMOTE is simple and effective, it struggles with sub-optimal performance with high dimensional datasets \parencite{maldonado2019alternative}, potential creation of noise samples and model overfitting \parencite{li2025improved}.

Generative adversarial networks (GAN) emerged as a framework to generate state-of-the-art realistic samples \parencite{goodfellow2014generative}. GANs do not need Markov chains, rather the gradients can be obtained only using backpropagation. In \parencite{mirza2014conditional} the authors extend GAN architecture to a conditional model where both generator and discriminator are conditioned on some kind of extra information (auxiliary information). Both GAN and conditional GAN models were experimented with image data. Building upon the idea of conditional GAN, Conditional Tabular GAN (CTGAN) which is another GAN based data augmentation method presents a benchmark framework to address data imbalance in tabular data by engaging conditional generator and training bu sampling \parencite{xu2019modeling}. CTGAN was designed with mode specific normalization to handle columns having complicated distributions. Using severely imbalanced Credit Card Fraud dataset \parencite{dal2015calibrating}, \textcite{fiore2019using} generated synthetic samples for underrepresented minority class and reported an improved sensitivity with a slight increase of false positives. As Credit Card Fraud dataset contains numerical features only, \textcite{engelmann2020conditional} presented conditional wasserstein GAN (cWGAN) based algorithm which is capable of handling both numerical and categorical features simultaneously. This algorithm focuses specially on target column for better downstream classification task. cWGAN was also found to be performing better with severe non-linear datasets. CTAB-GAN, a conditional table GAN architecture, is capable of addressing long tail issues and modeling diverse data types in imbalanced tabular data by encoding data through a conditional vector \parencite{zhao2021ctab}. The authors reported that CTAB-GAN does not only deliver better Machine Learning (ML) utilities and statistical similarity but also showed reasonable differential privacy. \textcite{nock2022generative} introduced Generative Trees (GT) with adversarial training between generator and discriminator similar to GAN. GT also includes tree (graph) structure with stochastic activations at the arcs and leaf dependent data generation. GTs have some advantages over neural networks in terms of missing data imputation, interpretability and training on various feature types. However, studies found that GAN based data generation algorithms are susceptible to membership inference attacks (MIA) and one of the main reasons for MIA in GAN based architecture is overfitting \parencite{hagestedt2019mbeacon, chen2020gan}.

Diffusion model is a fairly recent development in the field of data generative models which performed exceptionally well for image data. \textcite{kotelnikov2023tabddpm} presented TabDDPM, a denoising diffusion probabilistic model, capable of working with tabular data and any feature types. The study reports that TabDDPM is capable of outperforming alternative data generative approaches including GAN based and Variational Autoencoder (VAE) based models but lags behind SMOTE in terms of synthetic data quality. However, TabDDPM is advantageous in generating synthetic samples with differential privacy. Financial Tabular Diffusion (FinDiff) was designed to generate mixed type financial tabular data \parencite{sattarov2023findiff}. Utilizing embedding encoding and addressing mixed modality financial data challenges, FinDiff demonstrated promising results in terms of fidelity, privacy and utility. One of the most challenging problems researchers face while working with tabular data is encountering missing values. Missing values in tabular data can emerge from different reasons such as dangers and difficulties in collecting data, data not being recorded accidentally or considering data as irrelevant to be recorded \parencite{yoon2018personalized, emmanuel2021survey}. To alleviate this limitation, \textcite{ouyang2023missdiff} proposed a diffusion based framework termed as MissDiff which trains on data with missing values. MissDiff is a propitious method compared to other methods of handling missing values in tabular data. Alternative methods such as deletion and imputation have the possibility of leading to a reduction in data diversity or a biased performance \parencite{bertsimas2021simple}.

During the reverse process of diffusion model, gaussian noise is gradually transformed to synthetic data using score function. This score function is estimated through neural networks. Consequently, diffusion models take comparatively longer time to generate synthetic data with an iterative process \parencite{dhariwal2021diffusion}. Whereas GAN based models use a single forward pass of neural network to generate data of competitive quality \parencite{jolicoeur2021gotta}. Therefore the research on decreasing the data generation time with diffusion process has gained significant attention in recent times. \textcite{san2021noise} proposed diffusion model with a fewer number of denoising steps in order to reduce the data generation time. However, this method requires careful attention to find the optimal step size which can vary from model to model. For image data, \textcite{nichol2021improved} presented a framework to generate low resolution images within a limited time frame and then enhancing the resolution to generate high resolution synthetic image. In their work, \textcite{kim2022stasy} utilized self-paced learning (SPL) in score-based generative models (SGM) for tabular data generation. SPL is a training method inspired from human learning in which only a subset of data with minimal training loss is selected for training and the whole dataset is considered in a gradual manner. \textcite{lee2023codi} presented Codi, a contrasive learning method which can deal two types of variables (Discrete and Continuous) simultaneously. \textcite{zhang2023mixed} introduced TABSYN, another score based diffusion model, utilizes variational autoencoder (VAE) to generate synthetic tabular data in latent space. TABSYN works exceptionally well to generate high quality samples at a faster rate. Since diffusion models are heavily dependent on data, they show biases towards certain input. This is more common in text to image synthesis \parencite{schramowski2023safe}. Diffusion models for tabular data also suffers from these biases in training data. \textcite{yang2024balanced} introduced a tabular diffusion model which extends from label only conditioning to multivariate feature-level conditioning in order to increase fairness. The fairness in this work indicates the integration of sensitive guidance in the model by leveraging a U-net architecture in the reverse diffusion process.      

Conventionally, diffusion models use neural networks as Universal Function Approximators (UFA) and rely on Stochastic Differential Equations (SDE) to generate synthetic samples. \textcite{song2020score, song2020denoising} proposed to solve Ordinary Differential Equations (ODE) instead of SDE to generate data and reported that the removal of noise from data generation algorithm facilitates faster convergence. However, \textcite{jolicoeur2021gotta} concluded SDE performs better than ODE at a definite time frame. Moreover, there are other UFAs as well such as decision trees and more intricate tree based methods including Random Forests and GBTs \parencite{watt2020machine}. Since neural networks find it difficult to learn from irregular patterns of the target function, it is outperformed by GBT as UFA \parencite{grinsztajn2022tree}.

The use of latent space in the realm of diffusion process is quite recent but not entirely unusual. Latent space plays a crucial role to express a complex dataset in a simple yet efficient and meaningful way. The idea of latent space is originated from the work of \textcite{kingma2013auto} where the authors use encoder-decoder network to compress a dataset into a lower dimension and then sample from it. Tabular Variational Autoencoder or TVAE was proposed by \textcite{xu2019modeling} where the authors modified the variational autoencoder or VAE for tabular data by necessary pre-processing. Moreover, they modified the loss function to train two neural networks using evidence lower bound (ELBO) loss. \textcite{fonseca2023tabular} presented a comprehensive review of the different data augmentation method in latent space. Since Diffusion Models (DM) are related to likelihood-based models, they often require excessive computer resources to train the model. To alleviate this challenge, \textcite{rombach2022high} proposed Latent Diffusion Models (LDM) which utilizes autoencoder for latent space representation and trains diffusion model there for image synthesis. AutoDiff, a generative method which combines autoencoder and score-based diffusion Model reported by \textcite{suh2023autodiff}. AutoDiff performs well in terms of tabular data synthesis with privacy preservation. TABSYN \parencite{zhang2023mixed} generates synthetic data using score based diffusion process within a well crafted VAE. 

Summarizing the relevant studies we conclude that although there have been several works regarding tabular data generation, only a few of them actually consider imbalanced tabular data. Advanced data generation technique such as diffusion models have not been investigated profoundly for tabular data until very recently. However, the diffusion models used in those studies were score-based which, despite its great success in tabular data generation, suffers from multiple challenges such as computational inefficiency \parencite{li2024accelerating, jolicoeur2021gotta} and difficulties in accurately estimating score functions \parencite{song2020score}. GBT-based diffusion models have overcome those hurdles yet remain significantly underexplored in tabular data generation. One such way to reduce computational inefficiency is to implement latent space encoding and generate synthetic data in that encoded latent space. Although latent space has the potential to make tabular data generation faster and efficient, it has severely been underutilized. No tabular data generation method has applied Transformer as a latent space encoder. In terms of performance metrics, very few studies have reported privacy score, recall score and F-1 score altogether. What sets us apart is the fact that we address all these challenges and utilize state-of-the-art processes to make the tabular data generation process efficient and reliable which is supported by the metrics we report. A summary of related data generation methods for tabular data is listed in Table \ref{tab:1} to illustrate the uniqueness and unprecedented contribution of our work.

\begin{table}[htbp]
\centering
\caption{Comparison of Generative Models with Specific Contribution}
\label{tab:1}
\resizebox{\textwidth}{!}{%
\begin{tabular}{
    L{2cm}   
    L{1cm}   
    L{1.7cm} 
    L{1.2cm} 
    L{1cm} 
    L{1.6cm} 
    L{1cm} 
    L{1cm}   
    L{1cm}   
    L{1cm}   
}
\toprule

Generative Method & Tabular Data & Imbalanced Tabular Data & Diffusion & GBT-based Diffusion & Transformer & Latent Space & Privacy Score & Recall Score & F1 Score \\

\midrule

SMOTE \parencite{chawla2002smote} & \cmark & \cmark & \xmark & \xmark & \xmark & \xmark & \xmark & \xmark & \xmark \\

TVAE \parencite{xu2019modeling} & \cmark & \xmark & \xmark & \xmark & \xmark & \xmark & \xmark & \xmark & \cmark \\

CTGAN \parencite{xu2019modeling} & \cmark & \xmark & \xmark & \xmark & \xmark & \xmark & \xmark & \xmark & \cmark \\

WGAN \parencite{engelmann2020conditional} & \cmark & \cmark & \xmark & \xmark & \xmark & \xmark & \xmark & \cmark & \cmark \\

Ctab-gan \parencite{zhao2021ctab} & \cmark & \xmark & \xmark & \xmark & \xmark & \xmark & \cmark & \xmark & \cmark \\

Stasy \parencite{kim2022stasy} & \cmark & \xmark & \cmark & \xmark & \xmark & \xmark & \xmark & \xmark & \cmark \\

LDM \parencite{rombach2022high} & \xmark & \xmark & \cmark & \xmark & \xmark & \cmark & \xmark & \cmark & \xmark \\

TabDDPM \parencite{kotelnikov2023tabddpm} & \cmark & \xmark & \cmark & \xmark & \xmark & \xmark & \cmark & \cmark & \cmark \\

Codi \parencite{lee2023codi} & \cmark & \xmark & \cmark & \xmark & \xmark & \xmark & \xmark & \xmark & \cmark \\

Findiff \parencite{sattarov2023findiff} & \cmark & \xmark & \cmark & \xmark & \xmark & \xmark & \cmark & \xmark & \xmark \\

TABSYN \parencite{zhang2023mixed} & \cmark & \xmark & \cmark & \xmark & \xmark & \cmark & \cmark & \cmark & \xmark \\

Forest-Diffusion \parencite{jolicoeur2024generating} & \cmark & \xmark & \cmark & \cmark & \xmark & \xmark & \xmark & \xmark & \cmark \\

Fair-tab-diffusion \parencite{yang2024balanced} & \cmark & \xmark & \cmark & \xmark & \xmark & \xmark & \cmark & \xmark & \xmark \\

PCAForest (This paper) & \cmark & \cmark & \cmark & \cmark & \xmark & \cmark & \cmark & \cmark & \cmark \\

EmbedForest (This paper) & \cmark & \cmark & \cmark & \cmark & \xmark & \cmark & \cmark & \cmark & \cmark \\

Attention-\\Forest (This paper) & \cmark & \cmark & \cmark & \cmark & \cmark & \cmark & \cmark & \cmark & \cmark \\

\bottomrule
\end{tabular}
}
\end{table}

\section{{Methodology}}\label{sec3}

In this section, we describe principles of diffusion process and its variants which are relevant to this study; Gaussian Diffusion and GBT-based Diffusion. Later we introduce our three proposed generative models with detailed design: PCAForest, EmbedForest and AttentionForest.

\subsection{Preliminaries on Diffusion Models}\label{3.1} Denoising Diffusion Probabilistic Model or Diffusion Model in short is a probabilistic model inspired from non-equilibrium thermodynamics in which a Markov Chain is progressed gradually while adding Gaussian noise to a sample and this whole process is reversed later to denoise and synthesize new samples. 

\subsubsection{Diffusion Models}\label{3.1.1}

Diffusion models are described as likelihood based models consist of forward and reverse Markov process to deal with the data. The forward process is defined as \(
q(x_{1:T} \mid x_0) = \prod_{t=1}^{T} q(x_t \mid x_{t-1})
\). The forward process does the work of gradual noise addition to an initial sample \(
x_0
\) of the data distribution \(
q(x_0)
\). The added noise is sampled from \(q(x_t \mid x_{t-1})\) which is a predefined distributions and whose variances are \(
\{ \beta_1, \ldots, \beta_T \}
\). 
After that, a latent variable \(
x_T \sim q(x_T)
\) is gradually denoised through the reverse process defined as \(
p(x_{0:T}) = \prod_{t=1}^{T} p(x_{t-1} \mid x_t)
\) and new samples are generated from \(
q(x_0)
\). Generally unknown distributions \(
p(x_{t-1} \mid x_t)
\) are approximated by a neural network having parameters \(
\theta
\). The parameters \(
\theta
\) are learned through the optimization of a variational lower bound (VLB) defined as the equation 
\begin{equation}
\label{eq:1}    
\log q(x_0) \geq \mathbb{E}_{q(x_0)} \Bigg[ 
\underbrace{\log p_\theta(x_0 \mid x_1)}_{L_0}
- 
\underbrace{\mathrm{KL}(q(x_T \mid x_0) \| q(x_T))}_{L_T}
- 
\sum_{t=2}^{T} \underbrace{\mathrm{KL}(q(x_{t-1} \mid x_t, x_0) \| p_\theta(x_{t-1} \mid x_t))}_{L_t}
\Bigg]
\end{equation}

\subsubsection{Gaussian Diffusion Models}\label{3.1.2}

Operating in \(
x_t \in \mathbb{R}^n
\) which is a continuous space, the goal of the forward Markov process in Gaussian diffusion model is to gradually transform the complex unknown data distribution to a known gaussian distribution.This goal is accomplished by defining a forward noising process $q$. This process takes input of a data distribution $x_0 \sim q(x_0)$ and generates latents $x_1$ through $x_T$ with the addition of Gaussian noise at time $t$ with variance $\beta_t \in (0, 1)$.

\begin{equation}
\begin{aligned}
q(x_t \mid x_{t-1}) &:= \mathcal{N}\left(x_t ; \sqrt{1 - \beta_t} \, x_{t-1}, \, \beta_t \, \mathbf{I} \right) \\
q(x_T) &:= \mathcal{N}(x_T ; 0, \mathbf{I})
\end{aligned}
\label{eq:2}
\end{equation}

If the exact reverse distribution \( q(x_{t-1} \mid x_t) \) is known, the backward process of obtaining a sample from \( q(x_0) \) can be implemented through sampling from $x_T \sim \mathcal{N}(0, I)$. Since the reverse distribution $q(x_{t-1} \mid x_t)$ is affected by the complete data distribution, to estimate the distribution a neural network is used: 
\begin{equation}
p_\theta(x_{t-1} \mid x_t) := \mathcal{N}(x_{t-1}; \mu_\theta(x_t, t), \Sigma_\theta(x_t, t))
\label{eq:3}
\end{equation}

However, \textcite{ho2020denoising} recommended to use a diagonal \( \Sigma_\theta(x_t, t) \) with a constant \( \sigma_t \) and to compute \( \mu_\theta(x_t, t) \) as a function of \( x_t \) and \( \epsilon_\theta(x_t, t) \) in order to reduce the complexity of equation \eqref{eq:3}.
\begin{equation}
\mu_\theta(x_t, t) = \frac{1}{\sqrt{\alpha_t}} \left( x_t - \frac{\beta_t}{\sqrt{1 - \bar{\alpha}_t}} \, \epsilon_\theta(x_t, t) \right)
\label{eq:4}
\end{equation}
Here, \( \alpha_t := 1 - \beta_t,\quad\bar{\alpha}_t := \prod_{i \leq t} \alpha_i \) and \( \epsilon_\theta(x_t, t) \) does the work of predicting a noisy component \( \epsilon \) for \( x_t \). Finally, simplifying equation \eqref{eq:1}, 
\begin{equation}
\mathcal{L}_{\text{sim}}(\theta) = \mathbb{E}_{t,x_0,\epsilon}\!\left[\|\epsilon - \epsilon_\theta(\sqrt{\bar{\alpha}_t}x_0 + \sqrt{1-\bar{\alpha}_t}\,\epsilon, t)\|^2\right]
\label{eq:5}
\end{equation}


\subsubsection{Gradient Boosted Tree Diffusion Models}\label{3.1.3}

Decision tree is a process to determine the result of a function $f(x)$. This process implements a series of tests on the input \( x \) and the result of each tests decide the next test. This process is repeated until \( f(x) \) is determined with certainty \parencite{blockeel2023decision}. Each of the tests is considered as decision split. Decision tree maximizes predictive performance through the careful selection of decision splits given that it meets some certain criteria.

Gradient boosted trees or GBTs are an advanced form of decision tree in which each of the tree corrects the errors made by the previous step. The process of gradual correction starts from a simple tree called "weak learner" and progressively adds more trees with putting more emphasize on the instances on which the previous trees made incorrect prediction. This process does not stop until a certain level of accuracy is achieved. GBTs have been proven to be successful in tabular data prediction and classification \parencite{ma2020diagnostic}. XGBoost (Extreme Gradient Boosting) is one such GBT method which utilizes second order Taylor expansion, fast quantile splitting and other procedures to enhance its performance \parencite{jolicoeur2024generating, florek2023benchmarking}.  

Previous generative models such as GAN \textcite{goodfellow2014generative}, VAE \textcite{kingma2013auto} do not allow the use of GBTs due to the fact that the data generation process needs to pass through two differentiating models (Generator-Discriminator in GAN, Encoder-Decoder in VAE). Nevertheless, recently developed diffusion models use only one model which allows it to incorporate non-differentiable GBTs without any obstacle. 

The foundation of GBT-based approach lies in modeling the data generation process as a time-dependent transformation, where real data is progressively perturbed with Gaussian noise through a forward stochastic differential equation (SDE). The SDE in the model takes the form of the following equation,
\begin{equation}
dx = u_t(x)\,dt + g(t)\,dw
\label{eq:6}
\end{equation}

where, \( u_t(x) \) is a smooth, time-varying vector field, \( g(t) \) is a scalar function controlling the noise level, and \( w \) is Brownian motion. The objective of this transformation is to ensure that the data at time \( t = 0 \) is real, and by \( t = 1 \) it becomes nearly Gaussian noise. Furthermore, this process can be reversed and the reverse-time SDE allows the model to start from Gaussian noise and transform it back to realistic data.

This reverse process is simplified with the use of deterministic Ordinary Differential Equation (ODE) instead of SDE by conjecturing \( g(t) = 0 \) along with transforming the SDE into an ODE. This is done by assuming $g(t) = 0$, thereby turning the SDE into a standard ODE:
\begin{equation}
dx = v_\theta(t, x)\,dt
\label{eq:7}
\end{equation}

where, \( v_\theta(t, x) \) is a learned vector field approximated by the GBT model. This ODE delivers a sample from noise (at \( t = 1 \)) to realistic data (at \( t = 0 \)). 

To train the vector field \( v_\theta \), GBT-based diffusion \textcite{jolicoeur2024generating} utilizes Conditional Flow Matching (CFM) framework, particularly  I-CFM variant proposed by \textcite{tong2023improving}. The idea of this variant is to establish synthetic trajectories between pairs of real data points and noise samples, which eventually allows the model to learn the underlying flow field without requiring access to the true data distribution or its gradients. In order to initiate the process, for each real data point \( x_0 \), a noise vector \( x_1 \sim \mathcal{N}(0, I) \) is sampled and a conditional trajectory is defined as:
\begin{equation}
x(t) = t x_1 + (1 - t) x_0, \quad \text{for } t \in [0, 1]
\label{eq:8}
\end{equation}

This linear interpolation provides the input for the training. The following target vector field is defined as:
\begin{equation}
\mu_t(x \mid (x_0, x_1)) = x_1 - x_0
\label{eq:9}
\end{equation}

The model is then trained by minimizing the conditional flow matching loss:
\begin{equation}
\mathcal{L}_{\text{cfm}}(\theta) = \mathbb{E}_{t, x_0, x_1} \left\| v_\theta(t, x(t)) - (x_1 - x_0) \right\|^2
\label{eq:10}
\end{equation}

where, \( v_\theta \) is executed using an aggregation of GBT regressors. Each model represents a specific noise level $t$, and training is parallelized over different values of $t$ for computational efficiency.

\begin{figure}[!htbp]
  \centering
  \includegraphics[width=0.7\linewidth,height=0.55\textheight,keepaspectratio]{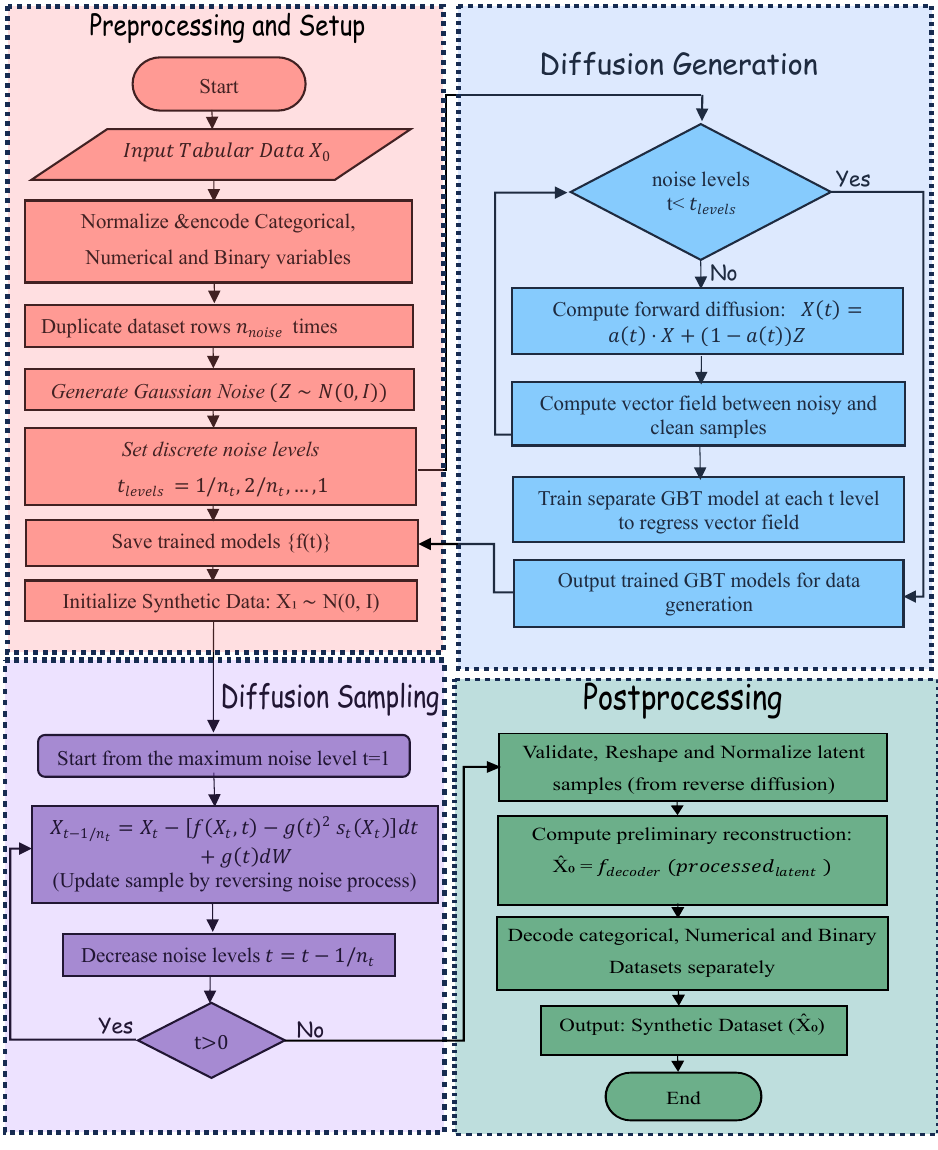}
  \caption{Synthetic Tabular Data Generation using Forest-Diffusion}
  \label{fig:your-label}
\end{figure}

\subsubsection{Autoencoder}\label{3.1.4}

An autoencoder is a type of neural network designed to learn a compressed representation of input data in an unsupervised manner. It consists of two key elements: an encoder, which reduces the dimensionality of the input, and a decoder, which attempts to reconstruct the original input from this reduced dimension. The primary goal is to capture the most relevant features of the data while minimizing the reconstruction error.

Given an input vector \( \mathbf{x} \in \mathbb{R}^n \), the encoder maps it to a latent representation \( \mathbf{z} \in \mathbb{R}^d \) using the transformation:
\begin{equation}
\mathbf{z} = h_\phi(\mathbf{x}) = \sigma(\mathbf{W}_e \mathbf{x} + \mathbf{b}_e)
\label{eq:11}
\end{equation}

where \( \mathbf{W}_e \) and \( \mathbf{b}_e \) are the parameters of the encoder that can be learned, and \( \sigma(\cdot) \) denotes a non-linear activation function such as ReLU or sigmoid. The resulting latent vector \( \mathbf{z} \) captures the essential characteristics of the input in a more compressed form.

The decoder reconstructs the input from the latent vector through a similar transformation:
\begin{equation}
\hat{\mathbf{x}} = g_\theta(\mathbf{z}) = \sigma(\mathbf{W}_d \mathbf{z} + \mathbf{b}_d)
\label{eq:12}
\end{equation}

Here, \( \mathbf{W}_d \) and \( \mathbf{b}_d \) are the weights and biases of the decoder. The output \( \hat{\mathbf{x}} \) is the reconstructed approximation of the original input \( \mathbf{x} \).

The network is trained by minimizing the reconstruction error across all training samples. A commonly used objective function is the Mean Squared Error (MSE), defined as:
\begin{equation}
\mathcal{L}_{\text{AE}} = \frac{1}{N} \sum_{i=1}^N \left\| \mathbf{x}^{(i)} - \hat{\mathbf{x}}^{(i)} \right\|^2
\label{eq:13}
\end{equation}

where $N$ is the number of training instances. This loss encourages the autoencoder to preserve the most informative aspects of the input while reducing noise and redundancy. The autoencoder learns a data representation that strikes a balance between compression and reconstruction accuracy, making it suitable for tasks such as dimensionality reduction, feature learning, and synthetic data generation.

\subsubsection{Transformer Autoencoder}\label{3.1.5}

Transformers are neural network architectures capable of processing sequential data exceptionally well by using self-attention mechanisms to estimate the importance of different parts of the input sequence \parencite{vaswani2017attention}. A Transformer fundamentally comprises of (i) multi head self attention layers to compute the dot product between the query, key, and value vectors for each element in the input sequence in order to weigh the importance of different parts of the input, (ii) feed forward neural networks to receive the output of self attention layers to make predictions and (iii) layer normalization to stabilize the training process by normalizing the inputs to each layer. A transformer autoencoder is a special type architecture that combines the strengths of transformer and autoencoder. In a transformer autoencoder, the encoder and decoder are both based on the transformer architecture where both components are built entirely upon self-attention mechanisms \parencite{cheng2024novel}. Given an input sequence 
\(
X = [x_1, x_2, \dots, x_n], \quad x_i \in \mathbb{R}^d
\)
the model first projects these inputs into a higher-dimensional embedding space using a learned linear transformation, followed by the addition of positional encoding to capture sequential information:
\begin{equation}
z_0 = XW_e + P
\label{eq:14}
\end{equation}
where, \( W_e \in \mathbb{R}^{d \times d_{\text{model}}} \) is the embedding matrix and \( P \in \mathbb{R}^{n \times d_{\text{model}}} \) denotes the positional encoding. The encoder then processes this sequence through multiple layers, each containing a multi-head self-attention mechanism and a position-wise feed-forward network. The self-attention mechanism is defined as
\begin{equation}
\text{Attention}(Q, K, V) = \text{softmax} \left( \frac{QK^\top}{\sqrt{d_k}} \right)V
\label{eq:15}
\end{equation}
where, \( Q, K, V \) are the query, key, and value matrices derived from the input. The multi-head version aggregates outputs from multiple attention heads:
\[
\text{MultiHead}(Q, K, V) = \text{Concat}(\text{head}_1, \dots, \text{head}_h)W^O
\]
with each head computed as 
\[
\text{head}_i = \text{Attention}(QW_i^Q, KW_i^K, VW_i^V)
\]
After passing through the encoder layers, the final hidden representation \( z \) serves as the latent code. The decoder, structurally similar to the encoder, uses masked self-attention to process the output sequence and includes an additional cross-attention mechanism over the encoder output to reconstruct the original input. The reconstruction \( \hat{X} \) is produced through these layers, and the model is optimized using a reconstruction loss—typically Mean Squared Error (MSE):
\begin{equation}
\mathcal{L}_{\text{rec}} = \frac{1}{n} \sum_{i=1}^{n} \left\| x_i - \hat{x}_i \right\|^2
\label{eq:16}
\end{equation}
This architecture offers an efficient way to capture both local and global dependencies in the input through self-attention, making it especially well-suited for tasks involving complex, structured or sequential data. Moreover, the attention mechanism enables more flexible and interpretable latent representations, which can enhance performance in downstream tasks such as reconstruction, classification, or anomaly detection.

\subsubsection{Principal Component Analysis (PCA)}\label{3.1.6}
PCA is a multivariate statistical method of dimensionality reduction. The idea of PCA is to retrieve
The most important information in the reduction of the size of the data and retain those information. \parencite{abdi2010principal}. The workflow of PCA is presented below:

\textbf{\textit{Data Standardization:}} The dataset is first standardized to ensure that each feature has zero mean and unit variance. This is essential because PCA is sensitive to the scale of the variables. Standardization is performed as:
\begin{equation}
z_{ij} = \frac{x_{ij} - \mu_j}{\sigma_j}
\label{eq:17}
\end{equation}
 where, \(x_{ij}\) is the value of the \(j\)-th feature for the \(i\)-th sample, \(\mu_j\) is the mean, and \(\sigma_j\) is the standard deviation of the \(j\)-th feature.

\textbf{\textit{Covariance Matrix Calculation:}} A covariance matrix is computed from the standardized data to determine the relationships between all pairs of variables. The covariance matrix $C$ is given by:
\begin{equation}
C = \frac{1}{n - 1} Z^{\top} Z
\label{eq:18}
\end{equation}
where, \( Z \) is the standardized data matrix and \( n \) is the number of observations.

\textbf{\textit{Eigenvalue Decomposition:}} The covariance matrix is decomposed into its eigenvalues and eigenvectors. The eigenvectors represent the principal components, and the corresponding eigenvalues indicate the amount of variance captured by each component. Mathematically:
\begin{equation}
C \mathbf{v}_i = \lambda_i \mathbf{v}_i
\label{eq:19}
\end{equation}
where, \( \lambda_i \) is the \( i \)-th eigenvalue and \( \mathbf{v}_i \) is the corresponding eigenvector.

\textbf{\textit{Principal Component Selection:}} The eigenvalues are sorted in descending order, and the top \(k\) eigenvectors corresponding to the largest eigenvalues are selected. These form the reduced-dimensional basis that retains the most significant variance.

\textbf{\textit{New Feature Space Projection:}} The original data is projected onto the selected principal components to obtain the transformed dataset in the reduced space. The projection is computed as:
\begin{equation}
X_{\text{PCA}} = ZW
\label{eq:20}
\end{equation}
where, \( W \) is the matrix formed by the selected \( k \) eigenvectors (principal components).

\subsection{Proposed Methods}\label{3.2}

We address the challenges of Diffusion Model and GBT-based Diffusion Model by formulating three different strategies. We develop frameworks to efficiently encode data to a latent space and leverage Diffusion Model in there. Our procedures differ from each other in the context of whether the encoded data has been decoded back to feature space or not. 

\subsubsection{PCAForest: PCA encoded Forest-Diffusion}\label{3.2.1}

Principal Component Analysis (PCA) has been widely used for dimensionality reduction considering it as a fast, reliable and efficient process to compress a large dataset and extract valuable insights from it. In this work, we incorporate PCA with Forest-Diffusion to compress the dataset into the latent space and train the GBT-based diffusion model in embedded space. 

\textbf{Step 1: Dataset Preparation and Standardization}
Imbalanced tabular dataset $\mathcal{D} = {X, y}$ is loaded, where $X \in \mathbb{R}^{n \times D}$ denotes feature matrix with $n$ and $D$ representing rows and columns respectively. This process is followed by standardization of the data through z-score $\tilde{X}_{ij}$ normalization using equation \eqref{eq:17}.

\textbf{Step 2: Principal Component Calculation}
This process of finding principal components starts with covariance matrix calculation followed by eigenvalue decomposition. 
\begin{equation}
\Sigma = \frac{1}{n} \tilde{X}^\top \tilde{X}
\label{eq:21}
\end{equation}

\begin{equation}
\Sigma = Q \Lambda Q^\top
\label{eq:22}
\end{equation}

Here, $\Sigma$ is the symmetric $n\times n$ matrix (covariance matrix), $Q$ is the matrix of eigenvectors, $\Lambda$ is a diagonal matrix of eigenvalues. Each column of $Q$ corresponds to a principal component (direction) and the diagonal values of $\Lambda$ indicate the variance explained by each component. 

Implementing the condition, $\frac{\sum_{i=1}^{d} \lambda_i}{\sum_{i=1}^{D} \lambda_i} \geq 0.95$, minimum $d$ is chosen so that 95\% or more variances are retained, where $d$ represents the number of principal components selected and $\lambda_i$ is the eigenvalue corresponding to the $i^{\text{th}}$ principal component. The selected principal components are projected according to the equation $X_\text{PCA} = \tilde{X} V_d$. 

\textbf{Step 3: Dimensionality Reduction}
By retaining only the top-$d$ principal components, $X_{\text{reduced}} \in \mathbb{R}^{n \times d}$, PCA reduces data dimensionality and makes it more tractable for the diffusion process. 

\textbf{Step 4: Train-Test Split}
$X_{\text{reduced}}$ is split into train and test in 7:3 ratio. 30\% test data is preserved for balanced dataset evaluation.
$\left( X_{\text{train}},\, y_{\text{train}} \right),\; \left( X_{\text{test}},\, y_{\text{test}} \right) = \text{StratifiedSplit}\left( X_{\text{reduced}},\, y,\, \text{70\% train} \right)$

\textbf{Step 5: Minority Class Identification}
Using label frequency $c_{\text{min}} = \arg\min_{c \in \{0, 1\}} \sum_{i=1}^{n} \mathbf{1}(y_i = c)$, minority class is identified and extracted. $X_{\text{minority}} = \{ x_i \mid y_i = c_{\text{min}} \}$

\textbf{Step 6: Synthetic Samples Generation}
At this phase, Forest-Diffusion Model (FDM) is employed and $\hat{n}$\% synthetic samples are generated where $\hat{n} \in \{25,\ 50,\ 75,\ \ldots,\ 300\}$.
\begin{equation}
\hat{X}_{\text{minority}} = \text{FDM}(X_{\text{minority}})
\label{eq:23}
\end{equation}

\textbf{Step 7: Data Augmentation}
Generated synthetic minority data $\hat{X}_{\text{minority}}$ is added to the original minority class to reduce the imbalance in the dataset. 
$X_{\text{aug}} = X_{\text{reduced}} \cup \hat{X}_{\text{minority}}, \quad
y_{\text{aug}} = y \cup \underbrace{\{c_{\text{min}}, \ldots, c_{\text{min}}\}}_{\hat{n} \text{ times}}$

\textbf{Step 8: Classifier Training on Balanced Data}
$X_{\text{aug}}$ is split into train and test set in 7:3 ratio and a classifier (Random Forest/XGBoost) is trained with 70\% train set. 
$\left( X'_{\text{train}},\, y'_{\text{train}} \right),\; \left( X'_{\text{test}},\, y'_{\text{test}} \right) = \text{StratifiedSplit}(X_{\text{aug}},\, y_{\text{aug}}, \text{70\% train})$
$f_{\text{aug}} = \text{Classifier}(X'_{\text{train}},\, y'_{\text{train}})$. We evaluate the model on \\$(X_{\text{test}}, y_{\text{test}})$ and report recall and F-1 scores.

Step 7 and Step 8 are repeated for every $\hat{n}$ in $\hat{n} \in \{25,\ 50,\ 75,\ \ldots,\ 300\}$.

\begin{algorithm}[htbp]
\caption{PCAForest}
\begin{algorithmic}[1]
\State \textbf{Input:} Dataset $\mathcal{D} = \{X, y\}$
\State \textbf{Output:} Trained classifier $f_{\text{aug}}$ and evaluation metrics

\State \textbf {Standardize}: $\tilde{X} := \text{StandardScaler}(X)$
    \State \textbf {PCA}: $X_{\text{PCA}} := \text{PCA}_d(\tilde{X})$ such that $\frac{\sum_{i=1}^{d} \lambda_i}{\sum_{i=1}^{D} \lambda_i} \geq 0.95$
\State \textbf {Split}: $(X_{\text{train}}, y_{\text{train}}), (X_{\text{test}}, y_{\text{test}}) := \text{StratifiedSplit}(X_{\text{PCA}}, y)$
\State \textbf {Minority sampling}: 
$c_{\min} := \arg\min_{c \in \mathcal{Y}} \text{count}(y = c)$, \quad
$X_{\min} := \{x_i \in X_{\text{PCA}} \mid y_i = c_{\min} \}$
\State \textbf {Synthesize}: $\hat{X}_{\min} := \text{FDM}(X_{\min})$
\State \textbf {Augment}: 
$X' := X_{\text{PCA}} \cup \hat{X}_{\min}$,\quad
$y' := y \cup \{c_{\min}\}^{|\hat{X}_{\min}|}$
\State \textbf {Retrain and evaluate}: 
$(X'_{\text{train}}, y'_{\text{train}}), (X'_{\text{test}}, y'_{\text{test}}) := \text{StratifiedSplit}(X', y')$\\
$f_{\text{aug}} := \text{Classifier}(X'_{\text{train}}, y'_{\text{train}})$\\
Evaluate $f_{\text{aug}}$ on original test set $(X_{\text{test}}, y_{\text{test}})$
\State \textbf {Compute evaluation metrics}: $\text{Recall}$ and $\text{F1 score}$
\end{algorithmic}
\end{algorithm}

\subsubsection{EmbedForest: Forest-Diffusion with Autoencoder}\label{3.2.2}

To further enhance data synthesis, we incorporate a feedforward Autoencoder to create a latent representation before diffusion. 

\textbf{Step 1: Encoding}
At the beginning of this process we use a neural network $f_\text{encoder}$ to map tabular data $X$ into a latent space.
\begin{equation}
Z_0 = f_{\text{encoder}}(X)
\label{eq:24}
\end{equation}

\textbf{Step 2: Latent Space Diffusion Process}
Gaussian noise is gradually added to the encoded data in forward diffusion process.
\begin{equation}
Z_t = \alpha_t Z_0 + \sqrt{1 - \alpha_t^2} \cdot \epsilon 
\label{eq:25}
\end{equation}

Here, $Z_0$ is the latent representation of the original data, $Z_t$ is the noised latent variable $\epsilon \sim \mathcal{N}(0, I)$ is the Gaussian noise and $\alpha_t$ is noise schedule scalar.

\textbf{Step 3: Training GBT to Predict Denoising Direction}
We define the training loss for our denoising field $v_\theta$ as a mean squared error between the predicted velocity and the true denoising direction, scaled appropriately according to the diffusion noise schedule.

\begin{equation}
\mathcal{L}_{\mathrm{reg}} = \mathbb{E}_{Z_0, \epsilon, t} \left[ 
\left\| v_\theta(Z_t, t) - \frac{Z_0 - Z_t}{\sqrt{1 - \alpha_t^2}} \right\|^2 
\right]
\label{eq:26}
\end{equation}
 
where, $Z_t = \alpha_t Z_0 + \sqrt{1 - \alpha_t^2} \cdot \epsilon$, $v_{\theta}(Z_t, t)$ is the vector field predicted by GBT, $\frac{Z_0 - Z_t}{\sqrt{1 - \alpha_t^2}}$ is the target which is derived by rearranging the forward noise equation to match the residual direction. 

\textbf{Step 4: Synthetic Data Generation by Reverse Diffusion}

\begin{equation}
\frac{dZ}{dt} = v_\theta(Z,t)
\label{eq:27}
\end{equation}

We use a deterministic reverse ODE to generate new samples. Starting with $Z_T \sim \mathcal{N}(0, I)$, we integrate equation \eqref{eq:27} backward from $t=T$ to $t=0$ and get denoised sample $Z_0$.

\textbf{Step 5: Decoding} Since the training and data generation process is performed in latent space, $Z_0$ is decoded back to original feature space. 
\begin{equation}
\hat{X}=f_{\text{decoder}}(Z_0)
\label{eq:28}
\end{equation}

Data augmentation, train-test split and evaluation process is implemented as described in \ref{3.2.1}.

\begin{algorithm}[htbp]
\caption{EmbedForest}
\begin{algorithmic}[1]
\State \textbf{Input:} Data $X$, noise schedule $\{\alpha_t\}$
\State \textbf{Output:} Synthetic sample $\hat{X}$
\vspace{2pt}

\State \textbf{Encode} data to latent space: $Z_0 = f_{\text{encoder}}(X)$
\State \textbf{Add noise:} $Z_t = \alpha_t Z_0 + \sqrt{1 - \alpha_t^2} \cdot \epsilon,\ \epsilon \sim \mathcal{N}(0, I)$
\State \textbf{Train GBT:} minimize
$
\mathcal{L}_{\mathrm{reg}} = \mathbb{E}_{Z_0, \epsilon, t} \left\| v_\theta(Z_t, t) - \frac{Z_0 - Z_t}{\sqrt{1 - \alpha_t^2}} \right\|^2
$
\State \textbf{Sample} $Z_T \sim \mathcal{N}(0, I)$
\State \textbf{Reverse ODE:} $\frac{dZ}{dt} = v_\theta(Z, t)$ from $t = T \rightarrow 0$ to obtain $Z_0$
\State \textbf{Decode:} $\hat{X} = f_{\text{decoder}}(Z_0)$
\end{algorithmic}
\end{algorithm}

\subsubsection{AttentionForest: Transformer encoded Forest-Diffusion}\label{3.2.3}
The ability of processing sequential data with the help of self attention mechanism makes a transformer-based Autoencoder an excellent tool to reduce the dimension and generate synthetic data in latent space. We use Transformer Autoencoder in our model which allows us to process a tabular data efficiently and generate high quality synthetic minority class data. The encoder component of a Transformer Autoencoder maps input tabular data into a
structured latent representation. Unlike conventional encoders that use fully connected layers, Transformer Autoencoders utilize a multi-head self-attention mechanism to dynamically weigh the
importance of different features. The workflow of AttentionForest is as follows:

\textbf{Step 1: Tokenization of Tabular Data} In AttentionForest encoder, each feature is considered as token. Since there are mainly two types of features, tokenization process is divided into two categories. In case of \textit{\textbf{Numerical Features}}, the data is projected into a continuous embedding space using a linear transformation. And for \textbf{\textit{Categorical Features}}, the data is converted into embeddings via an embedding layer. 

For a tabular input \(\mathbf{X} \in \mathbb{R}^{N \times d}\) where \(N\) is the number of samples and \(d\) is the feature dimension, the embedding space is defined as
\begin{equation}
E_i = W_i X_i + b_i, \quad i \in \{1, \ldots, d\}
\label{eq:29}
\end{equation}
Where, \(W_i\) and \(b_i\) are the learnable parameters for embedding transformation. 

\textbf{Step 2: Self Attention Mechanism} The unique design of transformer-based autoencoder is the center around the mechanism for self attention, which makes this model different from all other conventional approaches. The mathematical representation of this unique process is given in equation \eqref{eq:15}. It is applied to measure the significance of tokens in an input sequence with the purpose of understanding better the relations between them. Self-attention enables each feature in the tabular dataset to to pay attention to every other feature, capturing dependencies that would otherwise fail to be captured in traditional models.

\textbf{Step 3: Positional Encoding for Structured Data} The challenge with AttentionForest is that it lacks inherent ordering mechanism. Therefore, positional encoding are added to preserve the structured relationships among features:
\begin{equation}
PE(\text{pos}, 2i) = \sin\left( \frac{\text{pos}}{10000^{\frac{2i}{d}}} \right), \quad
PE(\text{pos}, 2i+1) = \cos\left( \frac{\text{pos}}{10000^{\frac{2i}{d}}} \right)
\label{eq:30}
\end{equation}
Where, \(pos\) is the position in the sequence $(0, 1, 2,....,T-1)$, $i$  is the dimension index(half sine, half cosine), d is the total and $2i$, $2i+1$ represent even and odd dimensions in the embedding vector. These encoding preserve structural information among features. 

\textbf{Step 4: Input Representation and Feature Encoding} After the completion of initial pre-processing, transformer autoencoder is utilized with GBT-based diffusion. The input tabular data is represented as $\mathbf{X}_0 \in \mathbb{R}^{N \times d}$ where \(N\) and \(d\) are number of samples and number of features respectively. Since there are two types of features, the processing method of each feature is different from other. \textbf{Categorical Features} are processed by applying an embedding layer. \textbf{Numerical Features} are processed through the application of a linear projection. 
\begin{equation}
\begin{aligned}
E_{\text{cat}} = \text{Embedding}(X_{\text{cat}}), \\
E_{\text{num}} = W_{\text{num}} X_{\text{num}} + b_{\text{num}}
\label{eq:31}
\end{aligned}
\end{equation}

Here, $\mathbf{E}_{\text{cat}} \in \mathbb{R}^{N \times d_{\text{emb}}}$ and $\mathbf{W}_{\text{num}} \in \mathbb{R}^{N \times d_{\text{emb}}}$.

\textbf{Step 5: Concatenation of Feature Embeddings and Transformer Encoding} 
All the feature embeddings are combined into a single matrix to initiate the latent space encoding.
\begin{equation}
E = [E_{\text{cat}}; \, E_{\text{num}}] \in \mathbb{R}^{N \times (d_{\text{cat}} + d_{\text{num}})}
\label{eq:32}
\end{equation}

The concatenated feature embeddings are passed through a AttentionForest encoder as defined in \ref{3.1.5},
\begin{equation}
L_0 = \text{TransformerEncoder}(E)
\label{eq:33}
\end{equation}
where \(L_0 \in \mathbb{R}^{N \times d_{\text{latent}}}\) is the latent representation.

\textbf{Step 6: Forward Diffusion in Latent Space} Once the input data is encoded into the latent space \(L_0\), forward diffusion is applied to
gradually add noise. This process starts with defining a noise schedule \(\alpha_t\) where \(t \in [0,1]\) controls the amount of noise added at each step. At each noise level \(t\), the noisy latent representation is computed.

\begin{equation}
L(t) = \alpha_t \cdot L_0 + \sqrt{(1 - \alpha_t^2} \cdot Z
\label{eq:34}
\end{equation}
where $Z \sim \mathcal{N}(0, 1)$ is the Gaussian noise.

\textbf{Step 7: Training GBT to Predict Denoising Direction} We define the training loss for our denoising field $v_\theta$ as a mean squared error between the predicted velocity and the true denoising direction, scaled appropriately according to the diffusion noise schedule.

\begin{equation}
\mathcal{L}_{\mathrm{reg}} = \mathbb{E}_{L_0, Z, t} \left[ 
\left\| v_\theta(L(t), t) - \frac{L_0 - L(t)}{\sqrt{1 - \alpha_t^2}} \right\|^2 
\right]
\label{eq:35}
\end{equation}

Where, $L(t) = \alpha_t \cdot L_0 + \sqrt{1 - \alpha_t^2} \cdot Z$, $v_{\theta}(L(t), t)$ is the vector field predicted by GBT, $\frac{L_0 - L(t)}{\sqrt{1 - \alpha_t^2}}$ is the target which is derived by rearranging the forward noise equation to match the residual direction.

\textbf{Step 8: Synthetic Data Generation by Reverse Diffusion} The latent space representation \(L_0\) is reconstructed again from the noisy latent representation \(L(t)\) through this backward process.  This process starts with $L(t) \sim \mathcal{N}(0, I)$, a complete noise. We integrate equation \eqref{eq:36} from $t=T$ to $t=0$ to get denoised $L_0$. 

\begin{equation}
\frac{dL}{dt} = v_\theta(L,t)
\label{eq:36}
\end{equation}

\textbf{Step 9: Decoding the Encoded Data} The latent space representation $L_0$ is decoded back into feature space by passing it through the decoder,

\begin{equation}
E = \mathrm{TransformerDecoder}(L_0)
\label{eq:37}
\end{equation}

where $$E \in \mathbb{R}^{N \times (d_{\text{cat}} + d_{\text{num}})}$$. For feature specific decoding, each of the features is passed through different functions: 

\begin{equation}
\begin{aligned}
X_{\text{cat}} = \mathrm{Softmax}(W_{\text{cat}} E_{\text{cat}} + b_{\text{cat}}), \\
X_{\text{num}} = W_{\text{num}} E_{\text{num}} + b_{\text{num}}
\label{eq:38}
\end{aligned}
\end{equation}

MSE loss is calculated for numerical feature decoding. Eventually the features are aggregated together to provide the terminal output. Data augmentation, train-test split and evaluation process is implemented as explained in \ref{3.2.1}.

\begin{equation}
X_0 = [\hat{X}_{\text{cat}};\; \hat{X}_{\text{num}}]
\label{eq:39}
\end{equation}
\begin{figure}[htbp]
  \centering
  \includegraphics[width=\linewidth]{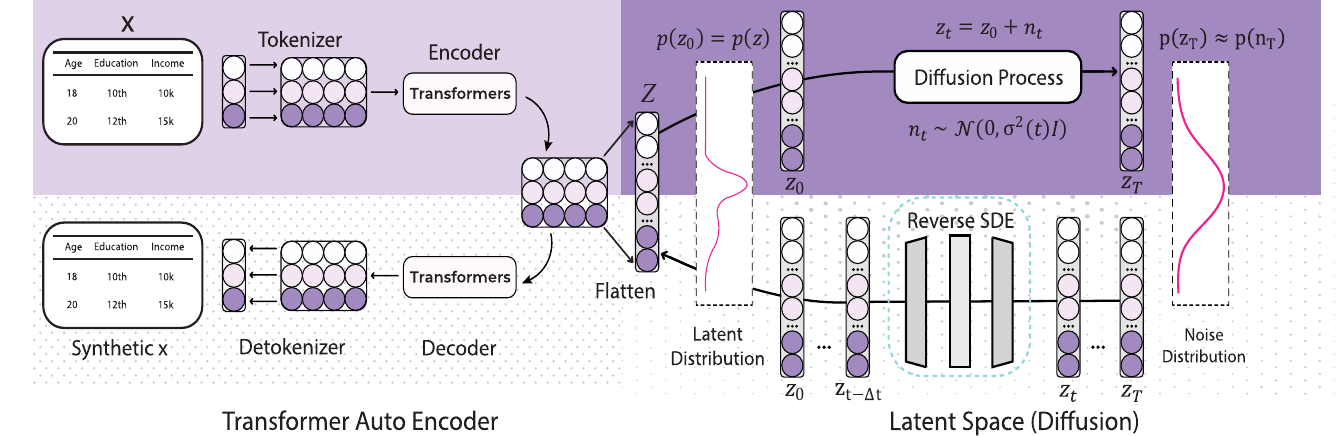}
  \caption{Flowchart for Latent Space Forest-Diffusion}
  \label{fig:your-label}
\end{figure}

\begin{algorithm}[htbp]
\caption{AttentionForest}
\begin{algorithmic}[1]
\State \textbf{Input:} Tabular dataset $\mathbf{X} \in \mathbb{R}^{N \times d}$
\State \textbf{Output:} Augmented dataset $\hat{\mathbf{X}}$

\State \textbf{Feature Tokenization:}
Project numerical features using linear layer, and categorical features using embedding layer:
\[
E_{\text{cat}} = \mathrm{Embedding}(X_{\text{cat}}), \quad
E_{\text{num}} = W X_{\text{num}} + b
\]

\State \textbf{Positional Encoding:}
Add sinusoidal positional encoding to preserve feature structure.

\State \textbf{Transformer Encoding:}
Concatenate embeddings and encode:
\[
E = [E_{\text{cat}}; E_{\text{num}}], \quad
L_0 = \mathrm{TransformerEncoder}(E)
\]

\State \textbf{Forward Diffusion:}
Apply noise using a predefined schedule:
\[
L(t) = \alpha_t L_0 + \sqrt{1 - \alpha_t^2} \cdot Z, \quad Z \sim \mathcal{N}(0, I)
\]

\State \textbf{Train GBT Denoiser:}
Minimize:
\[
\mathcal{L}_{\mathrm{reg}} = \mathbb{E}_{L_0, Z, t} \left\| v_\theta(L(t), t) - \frac{L_0 - L(t)}{\sqrt{1 - \alpha_t^2}} \right\|^2
\]

\State \textbf{Reverse Diffusion:}
Recover $L_0$ by solving $\frac{dL}{dt} = v_\theta(L, t)$ from noise.

\State \textbf{Decoding:}
Decode $L_0$:
\[
E = \mathrm{TransformerDecoder}(L_0)
\]
\[
\hat{X}_{\text{cat}} = \mathrm{Softmax}(W E_{\text{cat}} + b), \quad
\hat{X}_{\text{num}} = W E_{\text{num}} + b
\]

\State \textbf{Aggregation:}
Concatenate to form $\hat{\mathbf{X}} = [\hat{X}_{\text{cat}}; \hat{X}_{\text{num}}]$

\State \Return $\hat{\mathbf{X}}$
\end{algorithmic}
\end{algorithm}

\section{{Experimental Setup}}\label{sec4}
We design our GBT-based Diffusion model with a view to making this model accessible and feasible to work with for every the researchers in the field of tabular data. Tabular data itself has some intrinsic properties that makes it very difficult to work with. When this complicated tabular data is merged with Diffusion model, the data generation becomes an arduous task. Therefore, one of the fundamental goals of this study to make that process computationally efficient. We describe our experimental process here along with the datasets used in this study.   

\subsection{Experimental Environment}\label{4.1} We run our experiments for this study under ubuntu 22.04 on a machine equipped with 32 GB memory, a GeForce RTX 3070 GPU and a 8 core 16 thread AMD Ryzen 7 CPU. We use python programming language to give life to our mathematical expressions. We construct our PCAForest, EmbedForest and AttentionForest based on the work of \textcite{jolicoeur2024generating}. The hyperparameters used in this study are listed in Table \ref{tab:2}. We evaluate our data augmentation method on two machine learning models: Random Forest and XGBoost. We preserve 30\% of data exclusively for testing to avoid the biases.

\begin{table}[htbp]
\centering
\caption{List of Hyperparameter}
\label{tab:2}
\begin{tabular}{lc}

\toprule
Hyperparameter & Value \\
\midrule
Learning rate & 1 \\
Batch Size & 100 \\
Diffusion Timestep & 50 \\
Embedding Dimension         & 8  \\
Latent Dimension            & Data Dimension/2 \\
Number of Attention Heads   & 4 \\
Model Dimension             & 8 \\
Feedforward Hidden Dimension & 64 \\
Number of Transformer Layers & 2 \\
Batch Size                  & 32 \\
Number of Training Epochs    & 1000 \\
\bottomrule
    
\end{tabular}
\end{table}

\subsection{Datasets}\label{4.2} We use 11 real-world benchmark datasets with varying degrees of imbalance and originated from various fields including healthcare, power grid, environment, astronomy, sports, malware detection and fraud detection. Some of the datasets lacked sufficient imbalance and needed pre-processing to increase that. We employ one-hot encoding to categorical columns and remove rows with missing values before the start of data augmentation. Attributes such as name, SL No, ID are removed as well. The summary of the datasets are presented in Table \ref{tab:3}.  

\begin{table}
\centering
\caption{List of Datasets}
\label{tab:3}
\begin{tabular}{l c c c c r}

\toprule
Dataset & {\makecell[c]{Numerical\\Features}} & {\makecell[c]{Categorical\\Features}} & Sample Size & Imbalance Ratio & Source \\
\midrule
Mammography & 6 & 0 & 11183 & 42:1 & UCI \\
Pima Indians Diabetes & 8 & 0 & 560 & 8.33:1 & UCI \\
Smart Grid Stability & 12 & 0 & 7380 & 6.38:1 & Kaggle \\
Cardio Train & 5 & 6 & 39021 & 8.76:1 & Kaggle \\
Churn Modelling & 5 & 5 & 9998 & 3.91:1 & Kaggle \\
Performance Prediction & 19 & 0 & 1040 & 3.98:1 & Kaggle \\
Credit Card Fraud & 30 & 0 & 284807 & 577.88:1 & Kaggle \\
Oil & 39 & 10 & 937 & 21.85:1 & Kaggle \\
Malware Detection & 0 & 86 & 18157 & 4.15:1 & Kaggle \\
Spambase & 57 & 0 & 3401 & 4.55:1 & UCI \\
COIL-2000 & 1 & 84 & 9822 & 15.76:1 & UCI \\
\bottomrule
\end{tabular}
\end{table}

\subsection{Baselines}\label{4.3} Our data augmentation method is compared with four baseline methods of various types.

\textbf{SMOTE:} One of the earliest data augmentation techniques which is based on interpolation between two data points \parencite{chawla2002smote}. Due to its fast and efficient data generation, it has been used as baseline models for many studies.

\textbf{CTGAN:} Adaptation of GAN-based data generation method for tabular data \parencite{xu2019modeling}. The synthetic data is created by training two adversarial neural networks, generator and discriminator. 

\textbf{Forest-Diffusion:} The first work that shows GBT-based Diffusion that leverages gradient boosted trees in the data gaining process \parencite{jolicoeur2024generating}. The reverse diffusion process is accelerated by using the gradient boosted trees for this purpose.

\subsection{Evaluation Metrics}\label{4.4} The evaluation is carried out in three categories on our generative model. Machine Learning Utility and Statistical Similarity are used as quality measures for generated synthetic data and performance of downstream machine learning tasks, while Privacy Preservation metrics are used as a measure of the risk of identity breach.

\subsubsection{\textbf{Machine Learning Utility}}\label{4.4.1} We assess the performance of our generative models on downstream machine learning tasks by computing \textbf{Recall} and \textbf{F-1 Score}. Recall or True Positive Rate (TPR) is defined as the percentage of correctly classified positive outcomes out of all true positive instances.
\begin{equation}
\text{Recall} = \frac{\text{True Positives (TP)}}{\text{True Positives (TP)} + \text{False Negatives (FN)}}
\label{40}
\end{equation}

For an imbalanced dataset in binary classification, higher recall score is expected since it holds significant importance owing to the fact that an erroneous interpretation of a true positive case is more fatal than accidentally labeling a false case as true \parencite{shang2023precision}. Positive Predictive Value (PPV) or Precision is the ratio between true positive outcomes and total positive outcomes. In other words precision describes how many of the positive outcomes are correctly classified among all positively classified outcomes.
\begin{equation}
\text{Precision} = \frac{\text{True Positives (TP)}}{\text{True Positives (TP)} + \text{False Positives (FP)}}
\label{41}
\end{equation}

The harmonic mean of recall and precision is known as F-measure or F-1 score.
\begin{equation}
\text{F-1} = 2 \cdot \frac{\text{Precision} \cdot \text{Recall}}{\text{Precision} + \text{Recall}}
\label{42}
\end{equation}

F-1 tends to be closer to the smaller value. However, a higher value of F-1 indicates that both precision and recall score are reasonably high \parencite{miao2022precision}.

\subsubsection{\textbf{Statistical Similarity}}\label{4.4.2} \textbf{Wasserstein Distance (WD)} is a measurement method between two statistical distributions. WD can be interpreted informally as the cost to transport the mass of one distribution to the other \parencite{bomze2023optimization}.
For two probability distributions $P$ and $Q$ on a metric space $(\mathcal{X}, d)$, the Wasserstein distance is defined as:

\begin{equation}
WD(P, Q) = \inf_{\gamma \in \Pi(P,Q)} \, \mathbb{E}_{(x,y) \sim \gamma} \big[ d(x,y) \big]
\label{43}
\end{equation}

where, $\Pi(P,Q)$ is the set of all couplings (joint distributions) with marginals $P$ and $Q$. Unlike other distance metrics, WD can accommodate non-spherical distributions, and is also referred to as earth mover's distance. WD is estimated between real and synthetic data.\parencite{leo2023wasserstein}.

\subsubsection{\textbf{Privacy Evaluation}}\label{4.4.3} To inspect the privacy protection ability of original data, we utilize two distance metrics.

\textbf{Distance to Closest Record (DCR):} DCR is defined as the Euclidean distance between a synthetic data and its nearest neighbor in original data. For a point $x \in \mathcal{X}$ and a dataset $D = \{ x_1, x_2, \dots, x_n \}$ DCR is defined as:
\begin{equation}
d_{\min}(x, D) = \min_{x_i \in D, x_i \neq x} d(x, x_i)
\label{44}
\end{equation}

where, $d(x,x_i)$ = distance metric between $x$ and $x_i$. A DCR value of zero means that the produced synthetic data is perfectly resembling the original data and is more prone to leakage. On the other hand, a larger DCR value indicates a smaller identity breaching risk of the original data \parencite{liu2024scaling, zhao2021ctab}.

\textbf{Nearest Neighbor Distance Ratio (NNDR):}  NNDR is a classical privacy evaluation method based on the ratio of the Euclidean distance between the closest and second closest real neighbors and the corresponding synthetic values. For a point $x \in \mathcal{X}$ and a dataset $D = \{ x_1, x_2, \dots, x_n \}$ let $d_1$ and $d_2$ are the distances to closest and second closest neighbors. Mathematically,
\begin{equation}
\text{NNDR}(x) = \frac{d_1(x)}{d_2(x)}, \quad
d_1(x) = \min_{x_i \in D} d(x, x_i), \quad
d_2(x) = \min_{x_i \in D \setminus \{x_\text{NN}\}} d(x, x_i)
\label{45}
\end{equation}
where, $d(x,x_i)$ = distance metric between $x$ and $x_i$ and $x_{NN}$ is the nearest neighbor of $x$. NNDR is typically computed as a scalar in the range of [0,1] where a larger value indicates a larger amount of privacy \parencite{liu2024scaling, zhao2021ctab}.  

\subsection{Machine Learning Models}\label{4.5}
For downstream machine learning applications, we have used the Random Forest and Extreme Gradient Boosting or XGBoost classifiers. Both of the classifiers are decision tree-based ensemble machine learning algorithms. Random Forest consists of a large number of independent classifiers (decision trees) and contains three major hyper-parameters, node size, number of trees and number of features sampled. Each of the decision tree in the ensemble is constructed from data samples taken with replacement from training set (i.e., bootstrap sample). Random operation is injected in the build process with the selection of samples subset and feature subset to ensure the independence of each decision tree and improve classification accuracy \parencite{parmar2018review}. XGBoost creates a series of decision trees in a sequential manner, with each subsequent tree learning off of and rectifying the mistakes of its predecessors by minimizing a loss function through gradient descent. It has advanced regularization to prevent overfitting, runs in parallel to increase speed, and gracefully handles missing data. It meticulously selects optimal splits, trims redundant branches, and weights the branches to strike the right balance between bias and variance, leading to highly accurate predictions \parencite{chen2016xgboost}. The hyperparameters of Random Forest and XGBoost used in this study are listed in Table \ref{tab:4}.

\begin{table}[h!]
\centering
\caption{Comparison of Hyperparameters: Random Forest vs. XGBoost}
\label{tab:4}
\begin{tabular}{lclcl}
\toprule
\multicolumn{2}{c}{\textbf{Random Forest}} & & \multicolumn{2}{c}{\textbf{XGBoost}} \\
\cmidrule{1-2} \cmidrule{4-5}
\textbf{Hyperparameter} & \textbf{Value} & & \textbf{Hyperparameter} & \textbf{Value} \\
\midrule
random state      & 42      & & use label encoder  & False \\
n estimators      & 100     & & eval metric        & logloss \\
criterion         & gini    & & random state       & 42 \\
max depth         & None    & & n estimators       & 100 \\
min samples split & 2       & & learning rate      & 0.1 \\
min samples leaf  & 1       & & max depth          & 6 \\
max features      & sqrt    & & subsample          & 1 \\
bootstrap         & True    & & colsample\_bytree  & 1 \\
                  &         & & reg alpha          & 0 \\
                  &         & & reg lambda         & 1 \\
                  &         & & gamma              & 0 \\
                  &         & & min child weight   & 1 \\
                  &         & & booster            & gbtree \\
                  
\bottomrule
\end{tabular}
\end{table}

\subsection{Significance of Recall Score for Imbalanced Data}\label{4.6}
Recall score is defined as the number of correct results relative to the number of expected results \parencite{franti2023soft}. Considering Credit Card Fraud dataset where fraudulent transactions are labeled as minority and secured transactions are labeled as majority class, if the target is to correctly identify fraudulent transactions then a successful identification of a fraudulent transaction as fraudulent is considered True Positive. Conversely, an incorrect labeling of fraudulent transaction as secured is considered as False Negative. In case of a binary class dataset maximum number of true positives are expected due to the fact that if a fraudulent case is approved as a secured one this can cause more damage than the situation where a secured case is labeled as fraudulent. Similarly, for a cancer detection dataset, a cancer affected patient being diagnosed as healthy is worse than a healthy patient being diagnosed as cancerous. In consequence of the mentioned cases, a higher call gives the indication that the model is performing well to distinguish between positive and negative results. For this reason we provide much importance on Recall Score as machine learning utility metric.

\newpage

\section{Observations and Results}\label{sec5}
We evaluated the performance of our latent space encoded diffusion model methods across several metrics, including model-specific metrics like Recall and F1-score for Random Forest (RF) and XGBoost (XGB) classifiers, as well as synthetic data quality metrics such as the Wasserstein Distance (WD). To evaluate the privacy preservability we compute two widely used metrics: Nearest Neighbor Distance Ratio (NNDR) and Distance to Closest Record (DCR). Performance metrics vary across {\bfseries Augmentation Ratio}, the percentage of synthetic minority samples compared to the minority samples present in the base dataset, from 25\% to 300\%. For all tasks, we split the datasets in training(70\%) and testing(30\%) splits. The splitting was done prior to applying any data augmentation. This approach ensures that (i) the test set contains only original data samples, with no synthetic minority instances, thereby preserving its integrity for unbiased evaluation, and (ii) synthetic samples are utilized exclusively within the training set to enhance the model’s ability to recognize and classify minority class instances more effectively. 
 \begin{figure}[H]
    \centering
    \includegraphics[width=0.7\textwidth]{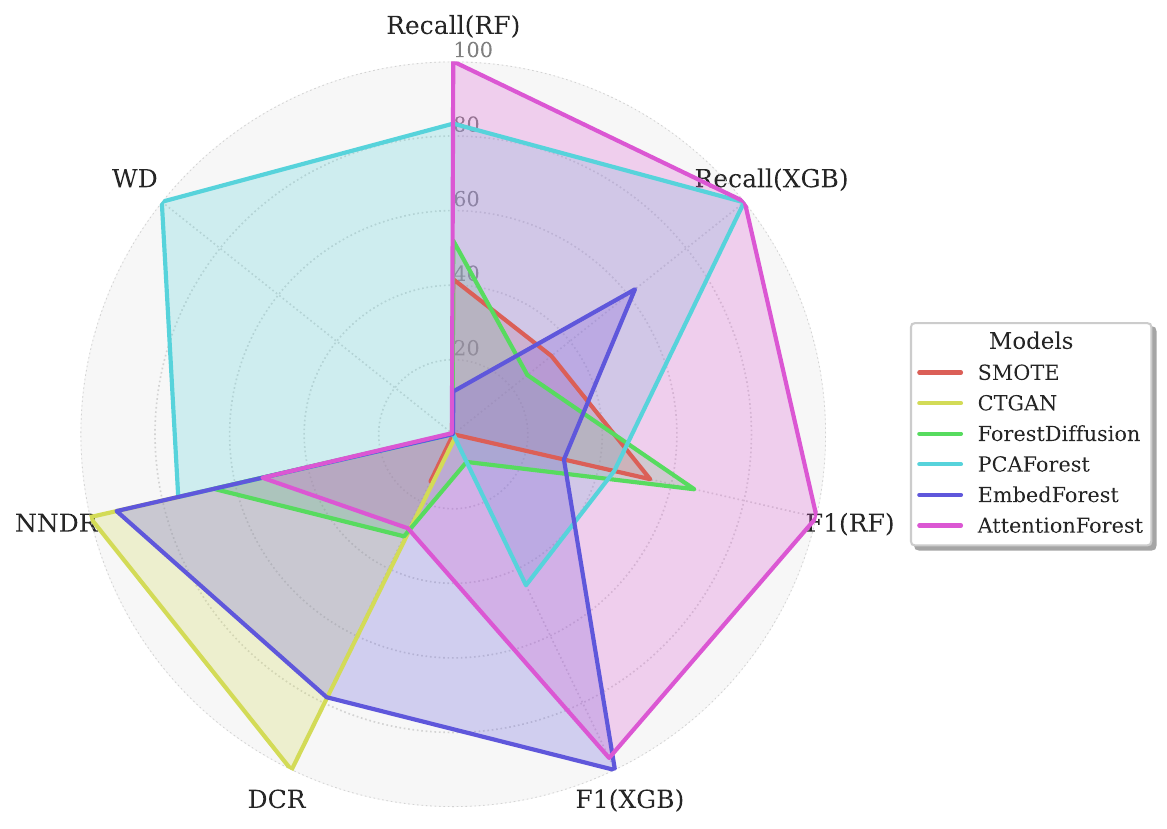}
    \caption{Our Latent Space Diffusion Models outperform Baseline Models in Downstream Machine Learning Tasks}
    \label{fig:radar}
\end{figure}
 As reflected in Figure \ref{fig:radar}, AttentionForest is not introducing noisy or overlapping samples, but instead preserves the decision boundaries and intrinsic data distribution. It shows the highest average Recall and F1 score for both Random Forest and XGBoost classifiers. The trade of between privacy and statistical similarity is also visible in Figure \ref{fig:radar}, as methods ensuring higher privacy lose statistical similarity from original data. PCAForest exhibits highest statistical similarity given that the WD evaluation is done in PCA embedded space. However, PCAForest achieves impressive recall scores both in Random Forest and XGBoost compared to baseline augmentation methods. On the other hand, EmbedForest does not manifest much improvement in recall scores but indicates higher privacy instead.  
\subsection{\textbf{Robustness in Minority Class Identification}\label{5.1}}
We have analyzed variation of recall scores for augmented data as augmentation ratio changes. Recall scores are measured for Random Forest and XGBoost classifier. 4 of our 11 datasets are listed in Tables \ref{tab:5}-\ref{tab:credit_card}, while all the other datasets show similar trends in terms of recall scores.
\begin{table}[htbp]
  \centering
  \caption{Recall Score Across Augmentation Ratios for Mammography Dataset}
  \label{tab:5}
  \resizebox{\textwidth}{!}{%
    \begin{tabular}{@{}c*{12}{c}@{}} 
      \toprule
      \multirow{2}{*}{\shortstack{Ratio}} & \multicolumn{2}{c}{SMOTE} & \multicolumn{2}{c}{CTGAN} & \multicolumn{2}{c}{Forest-Diffusion} & \multicolumn{2}{c}{PCAForest} & \multicolumn{2}{c}{EmbedForest} & \multicolumn{2}{c}{AttentionForest} \\
      \cmidrule(lr){2-3} \cmidrule(lr){4-5} \cmidrule(lr){6-7} \cmidrule(lr){8-9} \cmidrule(lr){10-11} \cmidrule(lr){12-13}
      & {\makecell[c]{Recall\\ (RF)}}& {\makecell[c]{Recall\\ (XGB)}} & {\makecell[c]{Recall\\ (RF)}}& {\makecell[c]{Recall\\ (XGB}}& {\makecell[c]{Recall\\ (RF)}}& {\makecell[c]{Recall\\ (XGB}} & {\makecell[c]{Recall\\ (RF)}}& {\makecell[c]{Recall\\ (XGB)}} & {\makecell[c]{Recall\\ (RF)}}& {\makecell[c]{Recall\\ (XGB)}} & {\makecell[c]{Recall\\ (RF)}}& {\makecell[c]{Recall\\(XGB)}} \\
      \midrule
      25  & 0.55 & 0.62 & 0.53 & 0.59 & 0.56 & 0.67 & 0.49 & 0.62 & 0.59 & \textbf{0.69} & 0.55 & \textbf{0.69} \\
      50  & 0.56 & 0.59 & 0.53 & 0.64 & 0.58 & 0.65 & 0.55 & 0.62 & 0.57 & 0.68 & \textbf{0.61} & \textbf{0.69} \\
      75  & 0.60 & 0.70 & 0.56 & 0.60 & 0.56 & 0.63 & 0.54 & 0.64 & 0.61 & 0.69 & \textbf{0.70} & \textbf{0.72} \\
      100 & 0.62 & 0.69 & 0.55 & 0.63 & 0.58 & 0.67 & 0.52 & 0.63 & 0.59 & 0.65 & \textbf{0.74} & \textbf{0.70} \\
      125 & 0.63 & 0.73 & 0.60 & 0.68 & 0.60 & 0.68 & 0.60 & 0.62 & 0.64 & 0.70 & \textbf{0.74} & \textbf{0.76} \\
      150 & 0.71 & 0.70 & 0.56 & 0.64 & 0.67 & 0.70 & 0.60 & 0.66 & 0.64 & 0.69 & \textbf{0.73} & \textbf{0.76} \\
      175 & 0.64 & 0.65 & 0.60 & 0.65 & 0.68 & \textbf{0.73} & 0.62 & 0.53 & 0.64 & 0.68 & \textbf{0.74} & 0.65 \\
      200 & 0.65 & 0.70 & 0.63 & 0.65 & 0.68 & 0.70 & 0.65 & 0.72 & 0.58 & 0.69 & \textbf{0.77} &\textbf{ 0.74} \\
      225 & 0.65 & 0.69 & 0.60 & 0.63 & 0.65 & 0.70 & 0.67 & 0.68 & 0.58 & \textbf{0.72} & \textbf{0.77} & 0.70 \\
      250 & 0.74 & 0.70 & 0.59 & 0.63 & 0.67 & \textbf{0.72} & 0.58 & 0.63 & 0.64 & 0.70 & \textbf{0.76} & 0.69 \\
      275 & 0.71 & 0.70 & 0.58 & 0.65 & 0.63 & 0.72 & 0.68 & 0.70 & 0.58 & 0.68 & \textbf{0.78} & \textbf{0.74} \\
      300 & 0.69 & 0.70 & 0.62 & 0.67 & 0.69 &\textbf{ 0.74} & 0.66 & 0.65 & 0.59 & 0.73 & \textbf{0.76} & \textbf{0.74} \\
      \bottomrule
    \end{tabular}
  }
\end{table}

It is clear that Our Novel methods show higher recall scores across most datasets and augmentation ratios. The EmbedForest Method shows less variation in performance across augmentation ratio. AttentionForest achieves higher recall score in general and when ratio increases, while PCAForest benefiting from a more sensitive classifier like XGBoost in latent space.

\begin{table}[htbp]
  \centering
  \caption{Recall Score Across Augmentation Ratios for Cardio Train Dataset}
  \label{tab:cardio}
  \resizebox{\textwidth}{!}{%
    \begin{tabular}{@{}c*{12}{c}@{}} 
      \toprule
      \multirow{2}{*}{\shortstack{Ratio}} & \multicolumn{2}{c}{SMOTE} & \multicolumn{2}{c}{CTGAN} & \multicolumn{2}{c}{Forest Diffusion} & \multicolumn{2}{c}{PCAForest} & \multicolumn{2}{c}{EmbedForest} & \multicolumn{2}{c}{AttentionForest} \\
      \cmidrule(lr){2-3} \cmidrule(lr){4-5} \cmidrule(lr){6-7} \cmidrule(lr){8-9} \cmidrule(lr){10-11} \cmidrule(lr){12-13}
      & {\makecell[c]{Recall\\ (RF)}}& {\makecell[c]{Recall\\ (XGB)}} & {\makecell[c]{Recall\\ (RF)}}& {\makecell[c]{Recall\\ (XGB)}}& {\makecell[c]{Recall\\ (RF)}}& {\makecell[c]{Recall\\ (XGB)}} & {\makecell[c]{Recall\\ (RF)}}& {\makecell[c]{Recall\\ (XGB)}} & {\makecell[c]{Recall\\ (RF)}}& {\makecell[c]{Recall\\ (XGB)}} & {\makecell[c]{Recall\\ (RF)}}& {\makecell[c]{Recall\\ (XGB)}} \\
      \midrule
      25  & 0.12 & 0.14 & 0.10 & 0.11 & 0.11 & 0.12 & 0.10 & \textbf{0.15} & 0.10 & 0.12 & \textbf{0.15} & 0.14 \\
      50  & 0.12 & 0.14 & 0.11 & 0.12 & 0.12 & 0.12 & 0.14 & \textbf{0.18} & 0.09 & 0.10 & \textbf{0.18} & 0.16 \\
      75  & 0.13 & 0.12 & 0.11 & 0.11 & 0.12 & 0.11 & 0.17 & \textbf{0.24} & 0.10 & 0.09 & \textbf{0.21} & 0.19 \\
      100 & 0.13 & 0.13 & 0.11 & 0.10 & 0.12 & 0.11 & 0.20 & \textbf{0.26} & 0.10 & 0.11 & \textbf{0.21} & 0.21 \\
      125 & 0.14 & 0.13 & 0.10 & 0.12 & 0.12 & 0.12 & 0.24 & \textbf{0.31} & 0.11 & 0.10 & \textbf{0.24} & 0.23 \\
      150 & 0.15 & 0.13 & 0.11 & 0.12 & 0.13 & 0.12 & 0.28 & \textbf{0.32} & 0.10 & 0.11 & \textbf{0.26} & 0.23 \\
      175 & 0.14 & 0.15 & 0.11 & 0.13 & 0.12 & 0.12 & 0.30 & \textbf{0.38} & 0.10 & 0.10 & \textbf{0.27} & 0.26 \\
      200 & 0.15 & 0.13 & 0.10 & 0.11 & 0.13 & 0.12 & 0.33 & \textbf{0.39} & 0.10 & 0.10 & \textbf{0.28} & 0.27 \\
      225 & 0.15 & 0.14 & 0.11 & 0.12 & 0.13 & 0.11 & 0.36 & \textbf{0.41} & 0.10 & 0.09 & \textbf{0.29} & 0.28 \\
      250 & 0.16 & 0.13 & 0.11 & 0.11 & 0.13 & 0.14 & 0.36 & \textbf{0.46} & 0.10 & 0.11 & \textbf{0.29} & 0.28 \\
      275 & 0.15 & 0.14 & 0.10 & 0.12 & 0.14 & 0.13 & 0.40 & \textbf{0.43} & 0.09 & 0.09 & \textbf{0.30} & 0.28 \\
      300 & 0.15 & 0.15 & 0.11 & 0.11 & 0.13 & 0.12 & 0.42 & \textbf{0.47} & 0.09 & 0.10 & \textbf{0.31} & 0.30 \\
      \bottomrule
    \end{tabular}
  }
\end{table}
In case of Churn Modelling Dataset(presented in Table \ref{tab:churn_modelling}, despite the majority of augmentation methods yielding comparable recall curves for XGBoost and Random Forest classifiers across different augmentation ratios, AttentionForest exhibits a gradual and continuous increase in recall with performance normally plateauing at the 200\% level of augmentation. PCAForest has more stepwise and abrupt improvement, though with effective performance. This is an indication of how attention-based mechanisms are insensitive to maintaining performance improvements but how dimension reduction-based mechanisms are unstable but occasionally extremely effective.

\begin{table}[htbp]
  \centering
  \caption{Recall Score Across Augmentation Ratios for Churn Modelling Dataset}
  \label{tab:churn_modelling}
  \resizebox{\textwidth}{!}{%
    \begin{tabular}{@{}c*{12}{c}@{}} 
      \toprule
      \multirow{2}{*}{\shortstack{Ratio}} & \multicolumn{2}{c}{SMOTE} & \multicolumn{2}{c}{CTGAN} & \multicolumn{2}{c}{Forest Diffusion} & \multicolumn{2}{c}{PCAForest} & \multicolumn{2}{c}{EmbedForest} & \multicolumn{2}{c}{AttentionForest} \\
      \cmidrule(lr){2-3} \cmidrule(lr){4-5} \cmidrule(lr){6-7} \cmidrule(lr){8-9} \cmidrule(lr){10-11} \cmidrule(lr){12-13}
      & {\makecell[c]{Recall\\ (RF)}}& {\makecell[c]{Recall\\ (XGB)}} & {\makecell[c]{Recall\\ (RF)}}& {\makecell[c]{Recall\\ (XGB)}}& {\makecell[c]{Recall\\ (RF)}}& {\makecell[c]{Recall\\ (XGB)}} & {\makecell[c]{Recall\\ (RF)}}& {\makecell[c]{Recall\\ (XGB)}} & {\makecell[c]{Recall\\ (RF)}}& {\makecell[c]{Recall\\ (XGB)}} & {\makecell[c]{Recall\\ (RF)}}& {\makecell[c]{Recall\\ (XGB)}} \\
      \midrule
      25  & 0.50 & 0.52 & 0.49 & 0.51 & 0.49 & 0.51 & 0.48 & 0.53 & 0.51 & 0.52 & \textbf{0.54} & \textbf{0.56} \\
      50  & 0.50 & 0.50 & 0.49 & 0.52 & 0.50 & 0.52 & 0.53 & 0.56 & 0.50 & 0.53 & \textbf{0.57} & \textbf{0.58} \\
      75  & 0.50 & 0.53 & 0.49 & 0.51 & 0.50 & 0.51 & 0.56 & 0.55 & 0.50 & 0.52 & \textbf{0.61} & \textbf{0.61} \\
      100 & 0.51 & 0.53 & 0.48 & 0.50 & 0.51 & 0.51 & 0.59 & 0.59 & 0.49 & 0.53 & \textbf{0.64} & \textbf{0.63} \\
      125 & 0.50 & 0.51 & 0.50 & 0.51 & 0.51 & 0.50 & 0.61 & 0.61 & 0.50 & 0.52 & \textbf{0.65} & \textbf{0.64} \\
      150 & 0.51 & 0.53 & 0.49 & 0.50 & 0.51 & 0.52 & 0.64 & 0.62 & 0.50 & 0.52 & \textbf{0.67} & \textbf{0.66} \\
      175 & 0.51 & 0.53 & 0.49 & 0.51 & 0.51 & 0.52 & 0.64 & 0.64 & 0.49 & 0.53 & \textbf{0.70} & \textbf{0.68} \\
      200 & 0.51 & 0.53 & 0.48 & 0.50 & 0.51 & 0.50 & 0.66 & 0.66 & 0.49 & 0.53 & \textbf{0.71} & \textbf{0.69} \\
      225 & 0.52 & 0.50 & 0.49 & 0.52 & 0.52 & 0.51 & 0.70 & 0.65 & 0.50 & 0.52 & \textbf{0.71} & \textbf{0.70} \\
      250 & 0.51 & 0.52 & 0.48 & 0.52 & 0.52 & 0.51 & 0.70 & 0.67 & 0.49 & 0.52 & \textbf{0.72} & \textbf{0.71} \\
      275 & 0.52 & 0.52 & 0.49 & 0.51 & 0.52 & 0.52 & 0.71 & 0.67 & 0.50 & 0.53 & \textbf{0.73} & \textbf{0.71} \\
      300 & 0.52 & 0.52 & 0.49 & 0.52 & 0.53 & 0.52 & 0.72 & 0.69 & 0.49 & 0.53 & \textbf{0.73} & \textbf{0.70} \\
      \bottomrule
    \end{tabular}
  }
\end{table}

The results for Credit Card Fraud dataset has the highest imbalance, with the majority class instances dominating the minority class instances by more than 577 times. This is reflected on the lack of increment in the recall score upon increaseing synthetic samples, shown in Table \ref{tab:credit_card}. even in this case, AttentionForest attains the highest scores among all the methods irrespective of augmentation methods. the improvement is shown in both Random Forest and XGBoost.

\begin{table}[htbp]
  \centering
  \caption{Recall Score Across Augmentation Ratios for Credit Card Fraud Dataset}
  \label{tab:credit_card}
  \resizebox{\textwidth}{!}{%
    \begin{tabular}{@{}c*{12}{c}@{}} 
      \toprule
      \multirow{2}{*}{\shortstack{Ratio}} & \multicolumn{2}{c}{SMOTE} & \multicolumn{2}{c}{CTGAN} & \multicolumn{2}{c}{Forest Diffusion} & \multicolumn{2}{c}{PCAForest} & \multicolumn{2}{c}{EmbedForest} & \multicolumn{2}{c}{AttentionForest} \\
      \cmidrule(lr){2-3} \cmidrule(lr){4-5} \cmidrule(lr){6-7} \cmidrule(lr){8-9} \cmidrule(lr){10-11} \cmidrule(lr){12-13}
      & {\makecell[c]{Recall\\ (RF)}}& {\makecell[c]{Recall\\ (XGB)}} & {\makecell[c]{Recall\\ (RF)}}& {\makecell[c]{Recall\\ (XGB)}}& {\makecell[c]{Recall\\ (RF)}}& {\makecell[c]{Recall\\ (XGB)}} & {\makecell[c]{Recall\\ (RF)}}& {\makecell[c]{Recall\\ (XGB)}} & {\makecell[c]{Recall\\ (RF)}}& {\makecell[c]{Recall\\ (XGB)}} & {\makecell[c]{Recall\\ (RF)}}& {\makecell[c]{Recall\\ (XGB)}} \\
      \midrule
      25  & 0.77 & 0.76 & 0.78 & 0.78 & 0.76 & 0.75 & 0.72 & 0.76 & 0.79 & 0.79 & \textbf{0.84} & \textbf{0.83} \\
      50  & 0.76 & 0.75 & 0.78 & 0.79 & 0.79 & 0.78 & 0.73 & 0.78 & 0.84 & 0.84 & \textbf{0.83} & \textbf{0.82} \\
      75  & 0.78 & 0.77 & 0.76 & 0.78 & 0.79 & 0.78 & 0.73 & 0.76 & 0.81 & 0.81 & \textbf{0.84} & \textbf{0.83} \\
      100 & 0.80 & 0.78 & 0.78 & 0.78 & 0.78 & 0.78 & 0.74 & 0.76 & 0.82 & 0.82 & \textbf{0.81} & \textbf{0.80} \\
      125 & 0.79 & 0.79 & 0.77 & 0.79 & 0.79 & 0.78 & 0.74 & 0.79 & 0.82 & 0.82 & \textbf{0.85} & \textbf{0.81} \\
      150 & 0.80 & 0.80 & 0.78 & 0.80 & 0.78 & 0.78 & 0.76 & 0.78 & 0.82 & 0.82 & \textbf{0.85} & \textbf{0.85} \\
      175 & 0.79 & 0.80 & 0.79 & 0.79 & 0.80 & 0.80 & 0.76 & 0.78 & 0.82 & 0.82 & \textbf{0.85} & \textbf{0.84} \\
      200 & 0.79 & 0.80 & 0.78 & 0.79 & 0.81 & 0.78 & 0.76 & 0.77 & 0.82 & 0.82 & \textbf{0.85} & \textbf{0.83} \\
      225 & 0.80 & 0.81 & 0.78 & 0.79 & 0.79 & 0.80 & 0.74 & 0.78 & 0.82 & 0.82 & \textbf{0.85} & \textbf{0.79} \\
      250 & 0.80 & 0.78 & 0.78 & 0.79 & 0.81 & 0.80 & 0.76 & 0.77 & 0.82 & 0.82 & \textbf{0.85} & \textbf{0.84} \\
      275 & 0.79 & 0.77 & 0.78 & 0.80 & 0.80 & 0.80 & 0.76 & 0.78 & 0.84 & 0.84 & \textbf{0.86} & \textbf{0.82} \\
      300 & 0.79 & 0.78 & 0.78 & 0.81 & 0.82 & 0.78 & 0.76 & 0.78 & 0.84 & 0.84 & \textbf{0.85} & \textbf{0.83} \\
      \bottomrule
    \end{tabular}
  }
\end{table}

\subsection{\textbf{Overall Performance Across Multiple Metrics}\label{5.2}}

We show the detailed performance of all models in Table \ref{tab:6},  presenting the best metric value per model (across augmentation ratios) for dataset. The experimental results show that our proposed methods, in particular AttentionForest, are always achieving better or quite competitive results in the majority of the datasets in terms of the important Recall and F1-Score metrics. However, it is noteworthy that, although our augmentation methods fall behind other methods in some cases, the trade-offs between privacy preservation and statistical similarity always gives an upper hand to our proposed methods. For example, on the Mammography dataset, AttentionForest attained a maximum Recall score of 0.78 and F1-Score of 0.79 which are significantly better than all the baseline models. A similar trend is seen with Smart Grid Stability dataset where AttentionForest again provided the highest F1-Score of 0.81. When implemented on the Churn Modelling dataset, AttentionForest's F1-Score of 0.64 is 3\% better than both SMOTE and CTGAN's score of 0.61. 

\small 
\begin{longtable}{>{\raggedright\arraybackslash}p{1.9cm}>{\raggedright\arraybackslash}p{2.3cm}*{4}{>{\centering\arraybackslash}p{1cm}}>{\centering\arraybackslash}p{1.3cm}*{2}{>{\centering\arraybackslash}p{1.2cm}}}
\caption{Performance Metrics Comparison Across Synthetic Data Generation Models and Datasets} \label{tab:6} \\

\toprule
Dataset & Model & \makecell{Recall\\(RF)} & \makecell{Recall\\(XGB)} & \makecell{F1\\(RF)} & \makecell{F1\\(XGB)} & DCR & NNDR & WD \\
\midrule
\endfirsthead

\multicolumn{9}{c}{Table \thetable{} -- Continued from previous page} \\
\toprule
Dataset & Model & \makecell{Recall\\(RF)} & \makecell{Recall\\(XGB)} & \makecell{F1\\(RF)} & \makecell{F1\\(XGB)} & DCR & NNDR & WD \\
\midrule
\endhead

\midrule
\multicolumn{9}{r}{\textit{Continued on next page}} \\
\endfoot

\bottomrule
\endlastfoot

\multirow{6}{*}{\makecell[l]{Churn\\Modelling}}
                & SMOTE & 0.52 & 0.53 & 0.60 & 0.61 & 870.82 & 0.49 & 668.92 \\
                & CTGAN & 0.50 & 0.52 & 0.60 & 0.61 & \textbf{6037.69} & 0.75 & 3456.09 \\
                & ForestDiffusion & 0.53 & 0.52 & 0.61 & 0.61 & 1899.16 & 0.66 & 842.91 \\
                & PCAForest & \textbf{0.72} & \textbf{0.69} & 0.61 & 0.60 & 1.26 & \textbf{0.85} & \textbf{0.09} \\
                & EmbedForest & 0.51 & 0.53 & 0.61 & 0.61 & 3884.34 & 0.74 & 1383.27 \\
                & AttentionForest & 0.73 & 0.71 & \textbf{0.64} & \textbf{0.62} & 1200.82 & 0.41 & 635.28 \\
\cmidrule{1-9}

\multirow{6}{*}{\makecell[l]{COIL-\\2000}} 
                & SMOTE & 0.07 & 0.08 & 0.10 & 0.12 & 2.26 & 0.50 & \textbf{0.17} \\
                & CTGAN & 0.07 & 0.07 & 0.10 & 0.11 & \textbf{14.26} & \textbf{0.97} & 0.46 \\
                & ForestDiffusion & \textbf{0.08} & 0.08 & \textbf{0.12} & 0.12 & 7.67 & 0.93 & 0.16 \\
                & PCAForest & 0.07 & \textbf{0.11} & 0.10 & 0.14 & 4.63 & 0.94 & 0.34 \\
                & EmbedForest & 0.06 & 0.09 & 0.09 & \textbf{0.14} & 8.48 & 0.94 & 0.45 \\
                & AttentionForest & 0.06 & 0.10 & 0.09 & 0.15 & 10.74 & 0.96 & 0.26 \\
\cmidrule{1-9}

\multirow{6}{*}{\makecell[l]{Mammo-\\graphy}}
                & SMOTE & 0.74 & 0.73 & 0.77 & 0.75 & 0.24 & \textbf{0.48} & \textbf{0.16} \\
                & CTGAN & 0.63 & 0.68 & 0.68 & 0.72 & \textbf{1.44} & 0.91 & 0.47 \\
                & ForestDiffusion & 0.69 & 0.74 & 0.73 & 0.74 & 0.48 & 0.61 & 0.18 \\
                & PCAForest & 0.68 & 0.72 & 0.68 & 0.70 & 0.39 & 0.62 & 0.25 \\
                & EmbedForest & 0.64 & 0.73 & 0.76 & \textbf{0.79} & 4.04 & \textbf{0.92} & 1.19 \\
                & AttentionForest & \textbf{0.78} & \textbf{0.76} & \textbf{0.80} & 0.77 & 0.55 & 0.50 & 0.18 \\
\cmidrule{1-9}

\multirow{6}{*}{\makecell[l]{Malware\\Detection}}
                & SMOTE & \textbf{0.93} & \textbf{0.94} & 0.93 & \textbf{0.94} & 0.07 & 0.06 & \textbf{0.01} \\
                & CTGAN & \textbf{0.93} & 0.93 & 0.93 & 0.93 & 1.80 & 0.96 & 0.03 \\
                & ForestDiffusion & \textbf{0.93} & 0.93 & \textbf{0.94} & 0.91 & 0.91 & 0.93 & 0.01 \\
                & PCAForest & 0.92 & 0.91 & 0.92 & 0.93 & \textbf{3.12} & 0.98 & 0.27 \\
                & EmbedForest & 0.91 & 0.93 & 0.93 & 0.94 & 0.72 & 0.99 & 0.04 \\
                & AttentionForest & 0.91 & 0.92 & 0.93 & 0.94 & 3.03 & \textbf{1.00} & 0.19 \\
\cmidrule{1-9}

\multirow{6}{*}{\makecell[l]{Performance\\Prediction}}         
                & SMOTE & 0.44 & 0.35 & 0.42 & 0.37 & 3.48 & 0.47 & 0.60 \\*
                & CTGAN & 0.25 & 0.32 & 0.34 & 0.42 & \textbf{16.86} & \textbf{0.92} & 2.35 \\*
                & ForestDiffusion & 0.43 & 0.32 & \textbf{0.44} & 0.38 & 7.53 & 0.77 & 0.80 \\*
                & PCAForest & \textbf{0.59} & \textbf{0.49} & 0.47 & 0.43 & 1.15 & 0.80 & \textbf{0.22} \\*
                & EmbedForest & 0.36 & 0.43 & 0.41 & \textbf{0.46} & 8.83 & 0.81 & 1.25 \\*
                & AttentionForest & 0.54 & 0.43 & 0.46 & 0.41 & 10.33 & 0.81 & 0.81 \\
\cmidrule{1-9}
\multirow[c]{6}{*}{\makecell[l]{Cardio\\ Train}}
                & SMOTE &        0.16 &         0.15 &    0.22 &     0.21 &    9.91 &  0.45 &   10.15 \\*
                & CTGAN &        0.11 &         0.12 &    0.17 &     0.19 &   38.29 &  0.83 &   98.65 \\*
                & ForestDiffusion &        0.14 &         0.14 &    0.20 &     0.20 &   27.51 &  0.81 &   15.83 \\*
                & PCAForest &        \textbf{0.42} &         \textbf{0.47} &    0.34 &     0.38 &    \textbf{0.59} &  0.81 &    \textbf{0.07} \\*
                & EmbedForest &        0.11 &         0.12 &    0.17 &     0.18 &  845.68 &  0.89 &  204.12 \\*
                & AttentionForest &        0.31 &         0.30 &    0.31 &     0.30 &   30.18 &  0.80 &   13.63 \\
                \cline{1-9}
                
\multirow[c]{6}{*}{\makecell[l]{Diabetes}}
                & SMOTE &        0.33 &         0.33 &    0.43 &     0.36 &   11.43 &  0.53 &    \textbf{6.51} \\
                & CTGAN &        0.33 &         0.39 &    0.36 &     0.41 &   60.50 &  0.90 &   16.68 \\*
                & ForestDiffusion &        0.33 &         0.22 &    0.40 &     0.30 &   16.89 &  0.57 &    6.84 \\*
                & PCAForest &        0.39 &         0.28 &    0.42 &     0.33 &    1.06 &  0.59 &    0.39 \\*
                & EmbedForest &        0.41 &         0.47 &    0.47 &     0.46 &   76.91 &  0.80 &   27.31 \\*
                & AttentionForest &        \textbf{0.47} &         0.35 &    \textbf{0.48} &     0.39 &   18.79 &  0.37 &   11.82 \\
                \cline{1-9}

\multirow[c]{6}{*}{\makecell[l]{Smart Grid\\Stability}} 
                & SMOTE &        0.62 &         0.76 &    0.70 &     0.76 &    \textbf{0.57} &  0.43 &    \textbf{0.11} \\*
                & CTGAN &        0.59 &         0.70 &    0.68 &     0.78 &    2.35 &  0.90 &    0.65 \\*
                & ForestDiffusion &        \textbf{0.70} &         \textbf{0.78} &    0.72 &     0.78 &    1.54 &  0.84 &    0.12 \\*
                & PCAForest &        0.68 &         0.74 &    0.67 &     0.73 &    1.77 &  0.87 &    0.13 \\*
                & EmbedForest &        0.59 &         0.77 &    0.72 &     \textbf{0.83} &    1.61 &  0.87 &    0.23 \\*
                & AttentionForest &        \textbf{0.70} &         0.76 &    \textbf{0.77} &     0.81 &    \textbf{0.91} &  0.45 &    0.10 \\
                \cline{1-9}

\multirow[c]{6}{*}{\makecell[l]{Spambase}} 
                & SMOTE &        0.88 &         0.90 &    0.90 &     0.91 &   16.98 &  0.54 &    2.92 \\*
                & CTGAN &        0.88 &         0.89 &    0.90 &     0.91 &  157.11 &  \textbf{0.92} &    7.62 \\*
                & ForestDiffusion &        0.86 &         0.90 &    0.90 &     0.91 &   27.67 &  0.78 &    3.22 \\*
                & PCAForest &        0.88 &         0.90 &    0.85 &     0.86 &    \textbf{3.37} &  0.94 &    \textbf{0.32} \\*
                & EmbedForest &        0.84 &         0.87 &    0.88 &     0.89 &   56.41 &  0.88 &    3.89 \\*
                & AttentionForest &        \textbf{0.85} &         0.86 &    0.88 &     0.89 &  437.02 &  0.99 &   18.65 \\
                \cline{1-9}

\multirow[c]{6}{*}{\makecell[l]{Oil}} 
                & SMOTE &        0.42 &         0.50 &    0.53 &     0.57 &  339349.84 &  0.61 &  21065.60 \\*
                & CTGAN &        0.33 &         0.33 &    0.47 &     0.40 & 1094477.00 &  0.89 &  52272.30 \\*
                & ForestDiffusion &        0.42 &         0.53 &    0.47 &     0.47 &  160824.33 &  0.62 &  26783.07 \\*
                & PCAForest &        0.42 &         \textbf{0.56} &    0.42 &     \textbf{0.56} &       \textbf{2.92} &  0.77 &      \textbf{0.47} \\*
                & EmbedForest &        \textbf{0.45} &         0.45 &    \textbf{0.56} &     0.56 & 8876449.25 &  0.82 & 235596.48 \\*
                & AttentionForest &        0.27 &         0.36 &    0.40 &     0.44 &   32448.97 &  0.65 &  12291.71 \\
                \cline{1-9}

\multirow[c]{6}{*}{\makecell[l]{Credit\\ Card\\Fraud}} 
                & SMOTE &        0.80 &         0.81 &    0.86 &     0.87 &     143.05 &  0.55 &    223.69 \\*
                & CTGAN &        0.79 &         0.81 &    0.86 &     0.86 &    4654.87 &  0.74 &   1121.21 \\*
                & ForestDiffusion &        0.82 &         0.80 &    \textbf{0.88} &     0.86 &     360.44 &  0.64 &    240.64 \\*
                & PCAForest &        0.76 &         0.79 &    0.84 &     0.86 &       \textbf{5.02} &  \textbf{0.90} &      \textbf{0.63} \\*
                & EmbedForest &        \textbf{0.84} &         \textbf{0.84} &    \textbf{0.88} &     \textbf{0.88} &     950.15 &  0.71 &    568.41 \\*
                & AttentionForest &        0.85 &         0.85 &    \textbf{0.88} &     0.87 &     123.10 &  0.55 &    132.61 \\
                
\end{longtable}

While we focus on classification performance, our models also obtain competitive scores in distribution similarity metrics like DCR, NNDR, and Wasserstein Distance (WD). In Mammography dataset, although the DCR value is highest for CTGAN but AttentionForest attains a comparable DCR and NNDR value with a better statistical similarity. In Credit Card Fraud dataset, AttentionForest achieves highest statistical similarity in data space without falling much behind ForestDiffusion in terms of privacy preservation. This trend is evident in other datasets as well which shows that our proposed methods synthesize not just good downstream task data but good synthetic data whose statistics make sense. Since PCAForest operates in embedded space the statistical similarity of this method is the highest in many of the dataset. However, AttentionForest outperforms all other methods in data space in a holistic comparison.

\begin{figure}[H]
    \centering
    \vspace{-12pt} 
    \includegraphics[width=\textwidth]{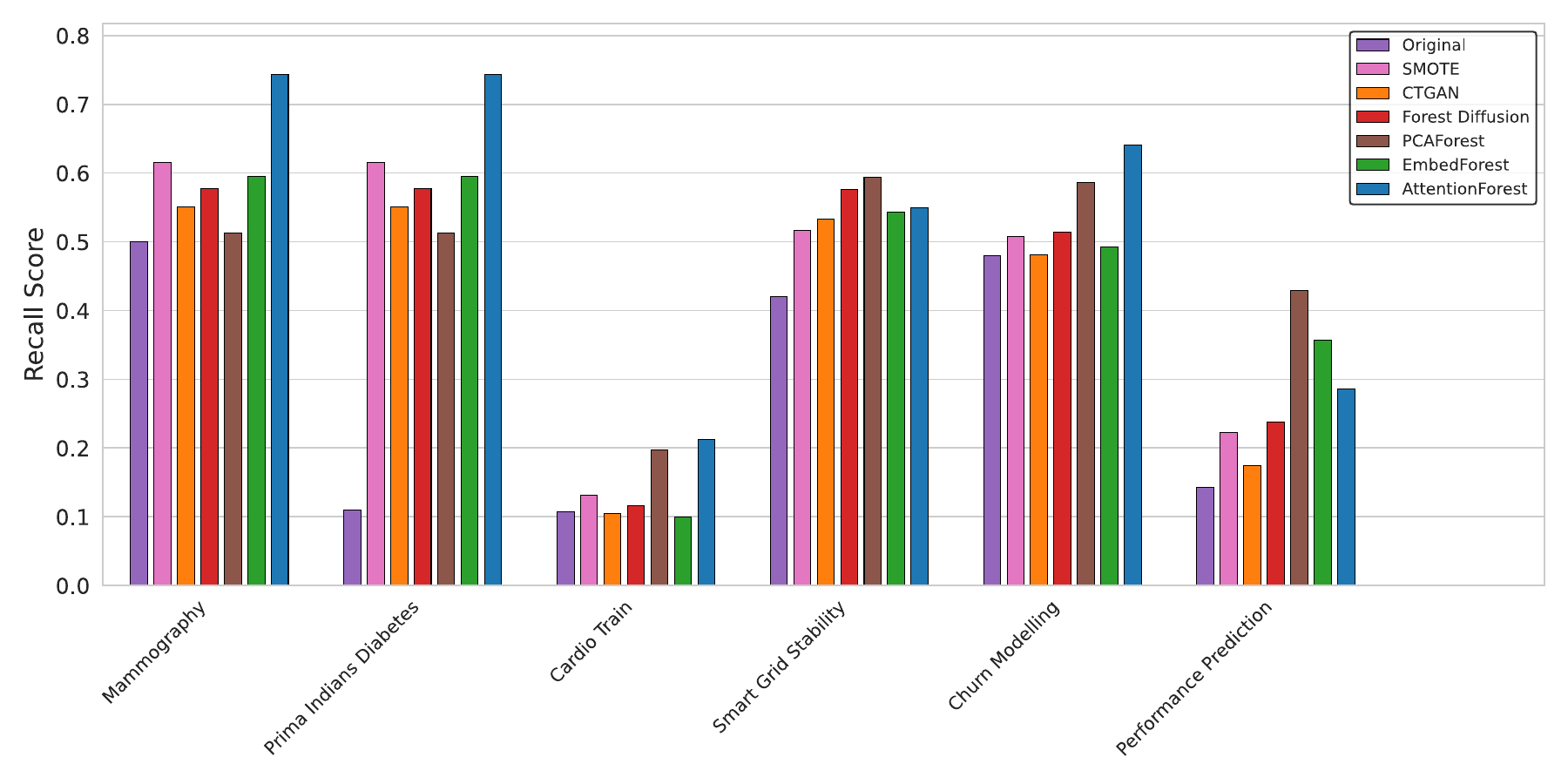}
    \vspace{-25pt} 
\end{figure}
\begin{figure}[H]
    \centering
    \vspace{-25pt} 
    \includegraphics[width=\textwidth]{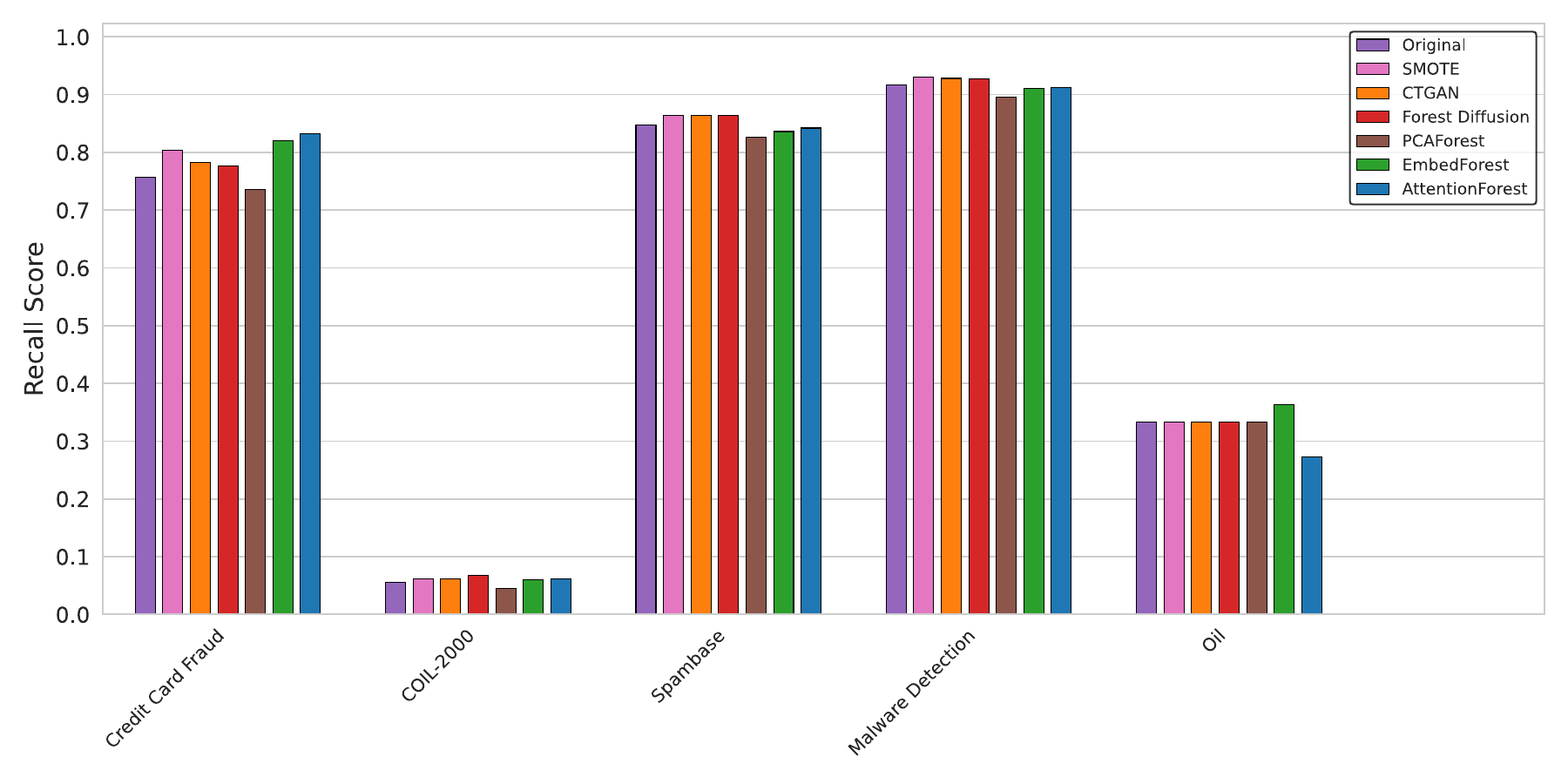}
    \caption{Recall Score Comparison at 100\% Augmentation Ratio Across Datasets}
    \label{fig:recall_com}
\end{figure}

 The bar plot in Figure \ref{fig:recall_com} shows a direct comparison between recall scores for a 100\% augmentation ratio. This visualization illustrates the practical advantages of our proposed methods on imbalanced data sets where correctly classification of the minority class is of primary importance.

The results indicate that our models produce a significant and clear performance gain. On the Mammography and Diabetes datasets, AttentionForest obtained a Recall score of around 0.75; this is much higher than the original, non-augmented data and the baseline methods. On the Churn Modelling dataset, all three of our proposed models (PCAForest, EmbedForest and AttentionForest) exceeded the established baselines. It should also be noted that on some datasets, for instance the COIL-2000 and Cardio Train datasets, the improvements are smaller. Although the differences among the top-performing models are marginal here, we find our proposed methods to be robust and competitive.

\subsection{\textbf{Impact of Augmentation Ratio on Performance Trends}}\label{5.3}
\begin{figure}[H]
    \centering
    \includegraphics[width=\textwidth]{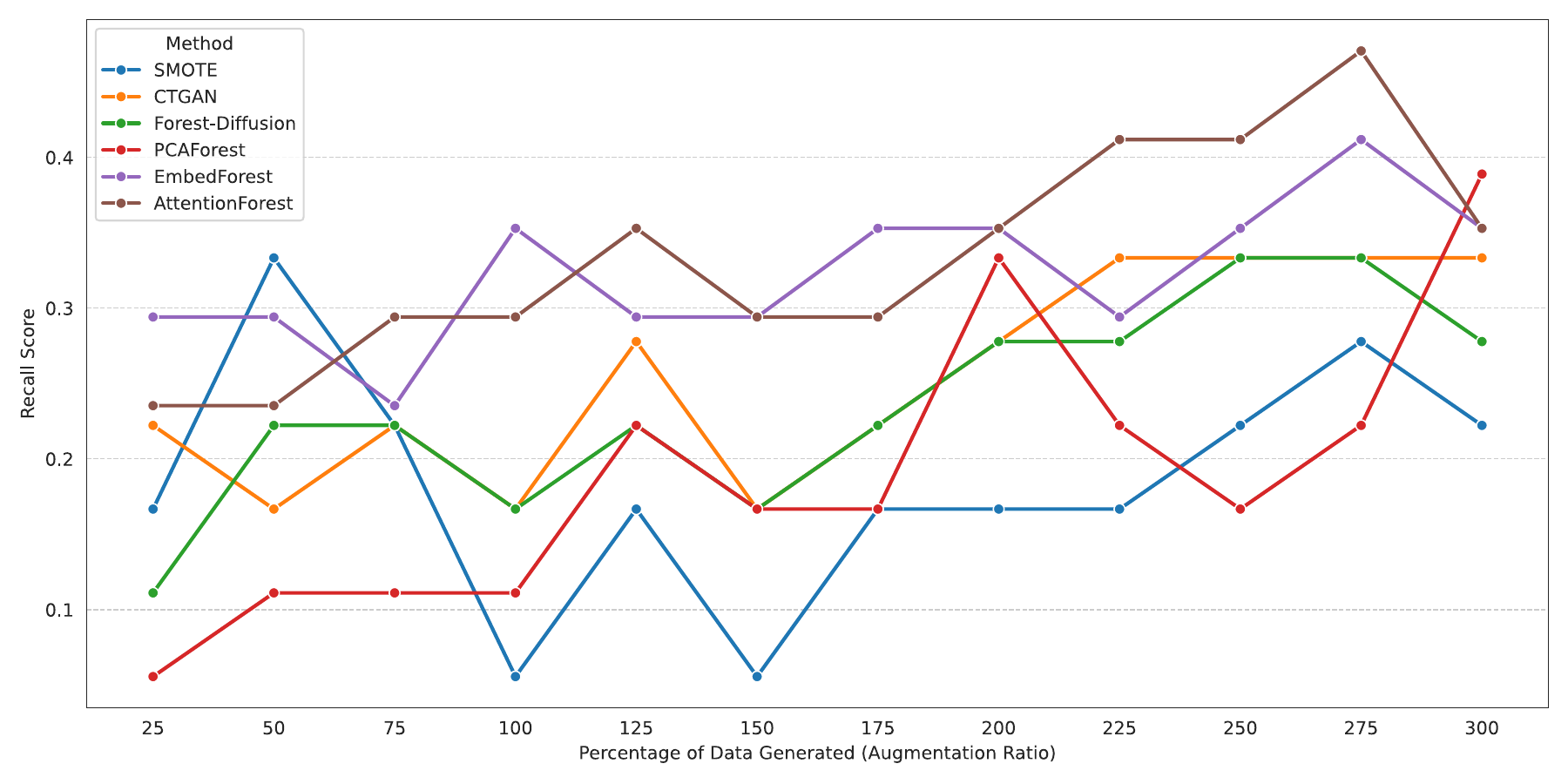}
    \caption{Recall Score Trends Across Augmentation Ratios}
    \label{fig:lineplot_of_ratio}
\end{figure}

The stability and robustness of the augmentation methods were tested by varying the amount of generated data from 25\% to 300\%. The resulting trends in Recall scores for Mammography dataset are plotted in Figure \ref{fig:lineplot_of_ratio}. Other datasets show similar trend with minor exception. This analysis reveals how different models behave under increasing levels of synthetic data injection.

Our proposed AttentionForest shows a good and generally stable upward trend with the increase in the augmentation ratio, becoming the leading model at higher ratios between 200\% and 275\%. This pathway is in sharp contrast with the volatility of baseline methods. For example, SMOTE has a very erratic behavior with a sharp peak at 50\% ratio and then quickly degrades. Moreover, CTGAN suffers in performance with an increase in augmentation levels. The better stability of our transformer-based method highlights that we have a reliable method for learning a large amount of high-quality synthetic data without any performance degradation.

\subsection{\textbf{Ablation study of AttenionForest Model}}\label{5.4}
To analytically assess the contribution of each architectural and training component to model performance, we performed a complete ablation study of our AttentionForest model. We implemented a baseline configuration and then separately changing key parameters in order to isolate the impact on Recall metric. Ablated are the parameters of the AttentionForest Autoencoder: embedding dimension (embed\_dim), num\_layers, nhead, latent\_dim and the parameters of flow based Forest-Diffusion model: n\_t, learning\_rate and duplicate\_K (for improved performance of forests).) and number of times to duplicate the data (for improved performance of forests) (duplicate\_K).

\begin{figure}[H]
    \centering
    \includegraphics[width=\textwidth]{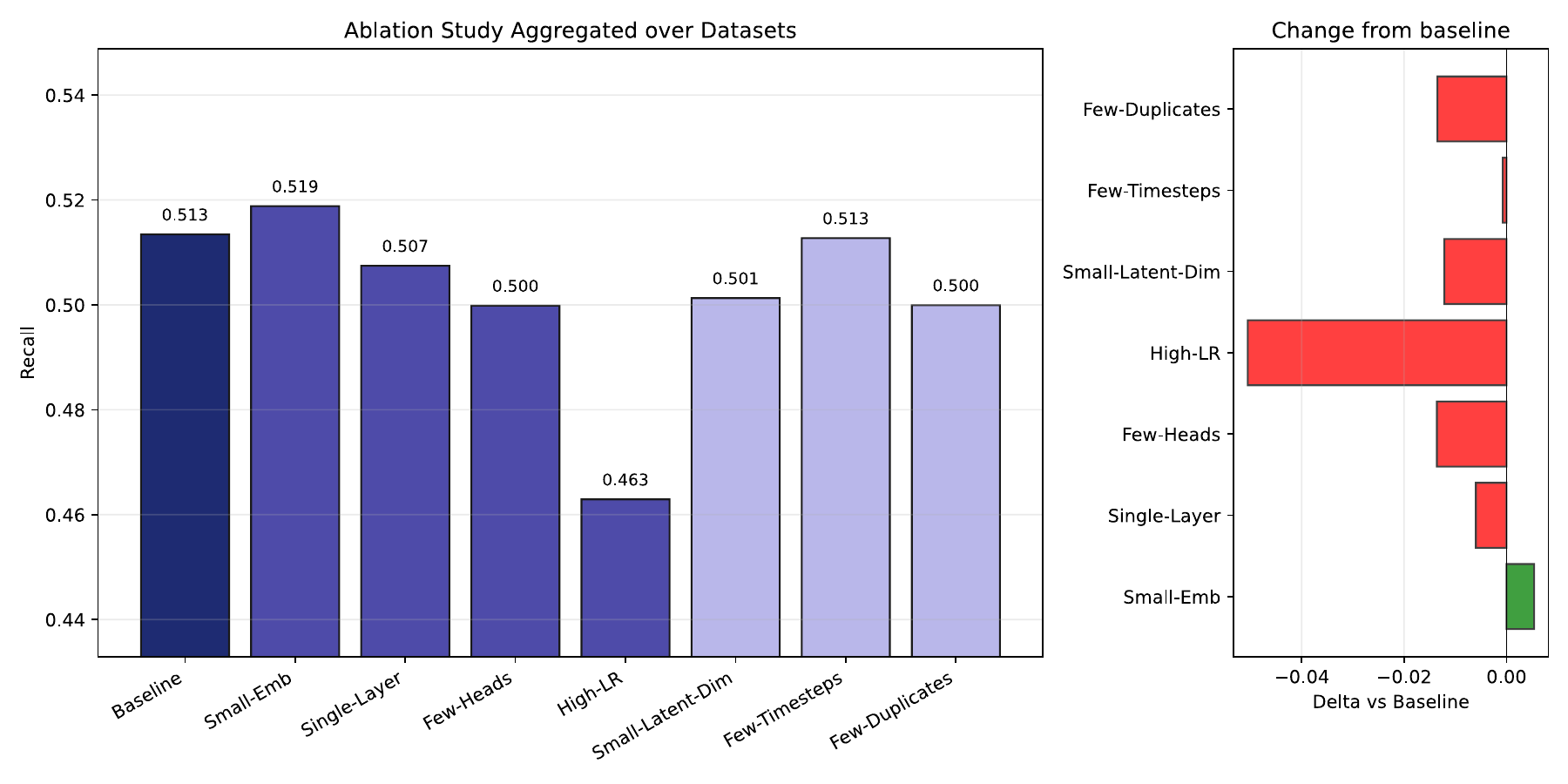}
    \caption{7 Parameter Ablations Study Aggregated Across Datasets}
    \label{img:agg_ablation}
\end{figure}

\begin{table}[h!]
\centering
\caption{Hyperparameter Configurations for Ablation Study}
\label{tab:ablation_configs}
\begin{tabular}{lcc}
\toprule
\textbf{Hyperparameter} & \textbf{Baseline Value} & \textbf{Ablation Values} \\
\midrule
\texttt{embed\_dim} & 16 & \{8, 16, 32\} \\
\texttt{num\_layers} & 2 & \{1, 2, 3\} \\
\texttt{nhead} & 4 & \{2, 4, 8\} \\
\texttt{latent\_dim\_factor} & 0.5 & \{0.25, 0.5, 1.0\} \\
\texttt{learning\_rate} & 0.001 & \{0.0001, 0.001, 0.01\} \\
\texttt{n\_t} & 50 & \{25, 50, 100\} \\
\texttt{duplicate\_K} & 100 & \{50, 100, 200\} \\
\bottomrule
\end{tabular}
\end{table}
Figure \ref{img:agg_ablation} presents the aggregated performance results averaged across all six datasets, providing a high-level overview of component sensitivity. The baseline model achieved a respectable average recall of 0.513. The analysis reveals that, on average, most modifications from the baseline either maintained or slightly degraded performance, underscoring the robustness of the initial configuration.
The significant average improvement was achieved when using a smaller embedding dimension (Small-Emb) which boosted the average recall to 0.519. This implies that a smaller feature representation might result in more generalisation across different tasks.

On the other hand, multiple modifications were found to be harmful to overall performance. The worst degradation was observed with a high learning rate (High-LR) where the average recall dropped by about 0.05 to 0.463. This indicates an acute sensitivity of the model to an aggressive learning rate that may prevent convergence. Similarly, diminishing the number of attention heads (Few-Heads) or using one transformer layer (Single-Layer) also had a negative effect on the average recall and lowered it to 0.500 and 0.507, respectively.


While the aggregated results provide a meaningful overall view, Figure 4 and Table 6 show that the ideal configuration depends very much on the characteristics of each dataset individually. In many cases, the optimum option for a particular component will vary widely depending on the task.
\begin{figure}[H]
    \centering
    \includegraphics[width=\textwidth]{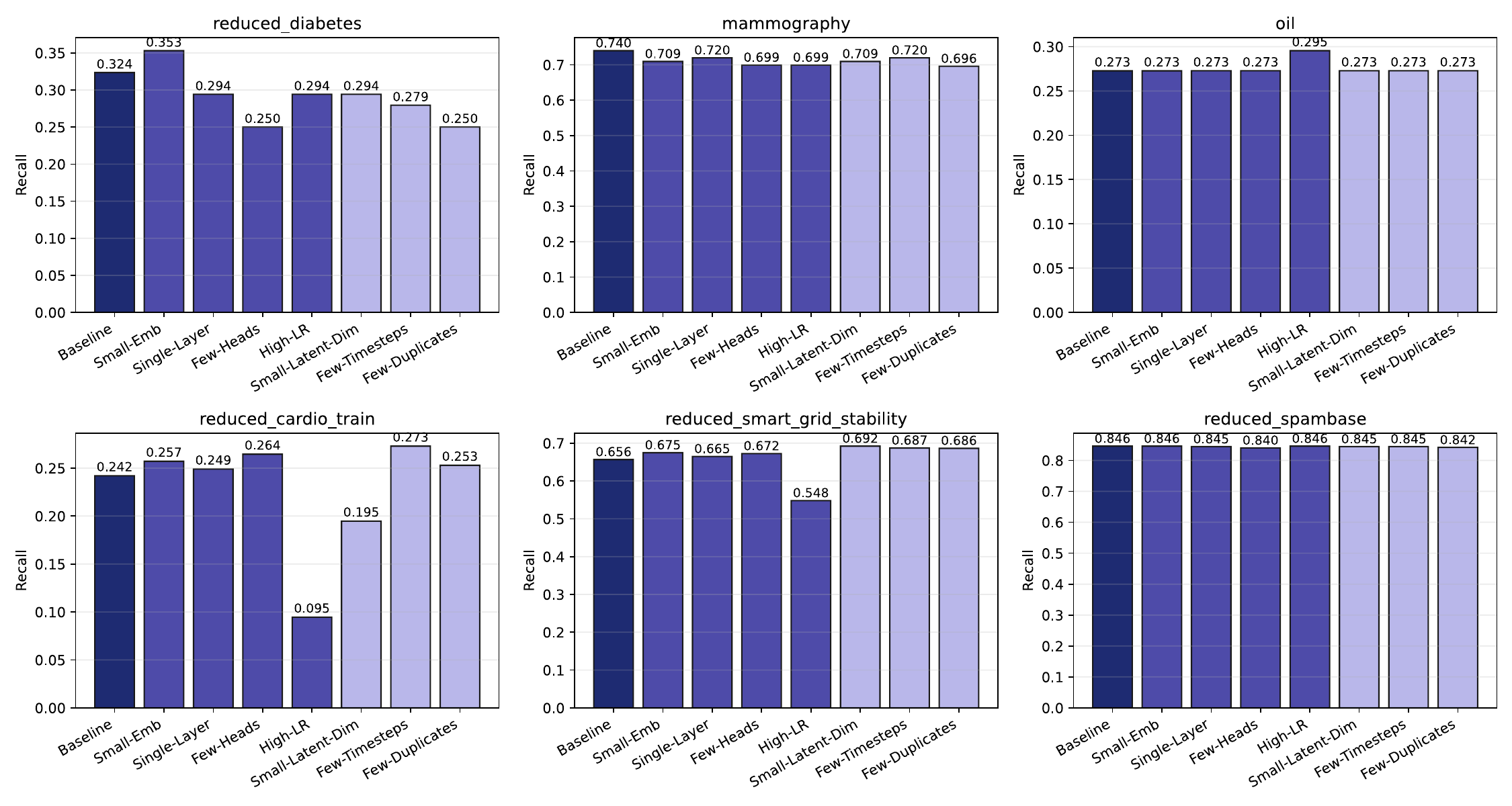}
    \caption{Ablation Study for 6 Datasets}
    \label{img:individual_ablation}
\end{figure}

For example, there is a clear task-dependency in the effect of embedding dimension. On the Prima Indian Diabetes dataset, an improved recall of 27.28\% was achieved with a smaller embedding dimension (embed\_dim = 8). In comparison, the Cardio Train dataset had the most to gain from using a higher embedding dimension (embed\_dim = 32), which improved recall by an impressive 37.84\%.

Similar data subset-specific significance was observed for other components. For the Oil dataset, increasing the number of attention heads (nhead) had the greatest positive effects on performance of - 33.35\%. For the Smart Grid Stability dataset, n\ t (number of diffusion time steps) was the most important hyperparameter that improved the recall by 20.55\%. The effect of the learning rate was also varied; although a higher learning rate was a bad idea on average, when the baseline performance was already very good, like on the Spambase dataset, a high learning rate had a small benefit of 2.17\% gain on that dataset. This is supported by Figure 7 which displays a catastrophic behavior of a high learning rate on the Cardio Train and Smart Grid Stability datasets that is clearly not neutral on Spambase.

\begin{table}[ht]
\centering
\caption{Ablation Study Results for Recall Across Various Datasets}
\label{tab:my_awesome_table}
\begin{tabular}{lclccc}
\toprule
Dataset & {\makecell[c]{Baseline\\ (Recall)}}  & {\makecell[c]{Best\\ Configured \\Component}} & {\makecell[c]{Best \\Component \\Value}} & {\makecell[c]{Recall\\Score for \\Component}} & {\makecell[c]{Improvement\\Over \\Baseline}} \\
\toprule
Prima Indian Diabetes & 0.323500 & embed\_dim & 8.000000 & 0.411765 & 27.28\% \\
Mammography & 0.739850 & num\_layers & 3.000000 & 0.783784 & 5.94\% \\
Oil & 0.272700 & nhead & 8.000000 & 0.363636 & 33.35\% \\
Cardio Train & 0.242025 & embed\_dim & 32.000000 & 0.333613 & 37.84\% \\
Smart Grid Stability & 0.656475 & n\_t & 100.000000 & 0.791391 & 20.55\% \\
Spambase & 0.846075 & learning\_rate & 0.010000 & 0.864407 & 2.17\% \\
\bottomrule
\end{tabular}
\end{table}

In summary, we establish two surprising findings from our ablation study: First, the baseline configuration is both robust and effective on a variety of tasks. Second, there is no optimum hyperparameter setting for all data. Finally, this work highlights the crucial need of conducting dataset-specific hyperparameter optimization (hyperparam tuning), most notably for sensitive parameters such as learning rate, embedding dimension and number of attention heads, to realize the full potential of the model and reach the best performance for a given task.

\subsection{\textbf{Maintaining Privacy without Losing Similarity}}\label{5.5}
One objective of our research was to investigate privacy of synthetic  minority class samples. We have attempted to maintain the privacy of synthetic minority class samples without significantly distancing them from the training data, as presented in Table \ref{tab:privacy}. Statistical methods of synthetic data generation like SMOTE (which uses KNN to determine synthetic samples), have low privacy (DCR=97.44, NNDR=0.46), as a result of having similarities with the original data in marginal and feature-wise distribution (WD=35.44). CTGAN has been shown to maintain the highest privacy (DCR=682.63, NNDR=0.88) at the cost of having lower similarity, as shown by a higher WD value of 352.43.   
\begin{table}[htbp]
\centering
  \caption{Value of Metrics Averaged over Datasets}
  \label{tab:privacy}
\scalebox{0.95}{
\begin{tabular}{lccccccc}
\toprule
Metric &  Recall(RF) &  Recall(XGB) &    F1(RF) &   F1(XGB) &         DCR &      NNDR &          WD \\
Model               &             &              &           &           &             &           &             \\
\midrule
SMOTE               &    0.39 &     0.419 &  0.460 &  0.45 &   97.44 &  0.45 &   35.43 \\
CTGAN               &    0.36 &     0.399 &  0.432 &  0.45 &  682.62 &  0.88 &  352.43 \\
ForestDiffusion     &    0.40 &     0.414 &  0.466 &  0.46 &  208.70 &  0.73 &   60.08 \\
PCAForest               &    0.43 &     0.460 &  0.455 &  0.47 &    1.79 &  0.77 &    0.16 \\
EmbedForest      &    0.37 &     0.437 &  0.448 &  0.49 &  534.88 &  0.85 &  165.73 \\
AttentionForest &    0.44 &     0.460 &  0.484 &  0.48 &  192.47 &  0.68 &   33.24 \\
\bottomrule
\end{tabular}
}
\end{table}
 Forest-Diffusion has Much higher DCR and NNDR values (208.71 and 0.73, respectively), indicating higher privacy than SMOTE while also having a higher WD value of 60.08.  Our transformer encoded latent space helps maintain a similar label of privacy to regular Forest (DCR=192.47, NNDR=0.68) with a WD value of 33.28, which is close to SMOTE. On the other hand, the latent space of the feedforward autoencoder architecture increases the DCR to 534.88 and NNDR to 0.85, which is closer to CTGAN. The WD value of 165.73, while higher than Forest-Diffusion and AttentionForest, remains much lower than CTGAN. as the PCAForest method keeps the augmented data in latent space, both the distribution privacy and distance remain low (DCR=1.80, NNDR=0.78, WD=0.16

\subsection{\textbf{Computational Efficiency Analysis}}\label{5.7}
We train flow-based diffusion models for synthetic data generation in latent space, where the dimension reduction accounts for reduced training time for the diffusion model, while some portion of the time is spent in the encode and decode period. As shown in Table \ref{tab:10}, substantial discrepancies exist in the computational demands of the evaluated methods which depend on (i)Nature and feature types of each dataset and (ii)architecture of each proposed methods.
The elevated runtime of AttentionForest is a direct consequence of its architectural design. The model integrates a transformer-based autoencoder with the diffusion process, necessitating an encoding-decoding cycle before applying Forest-Diffusion.  This cycle is computationally expensive due to the use of self-attention mechanisms inside transformers that require the processing of length sequences of features, scaling quadratically with input length. However, the time complexity of the model greatly exceeds other methods, especially in large scale datasets (COIL-2000: 1867.2 seconds, Malware Detection: 8274.9 seconds). While this complexity may yield more expressive latent representations, it introduces a considerable computational burden for larger datasets.

\begin{table}[htbp] 
\centering
\caption{Comparison of Data Augmentation Time (in seconds) at 100\% Augmentation Ratio} 
\label{tab:10}
\scalebox{0.95}{
\begin{tabular}{lcccc} 
\toprule \textbf{Dataset} & \textbf{Forest\_Diffusion} & \textbf{PCAForest} & \textbf{EmbedForest} & \textbf{\makecell[c]{Attention\\Forest}} \\ 
\midrule 
Diabetes & 19.5s & 16.1s & 11.1s & 20.3s \\ 
Cardio Train & 483.4s & 435.2s & 440.1s & 1054.8s \\ 
Mammography & 15.5s & 11.1s & 28.6s & 45.5s \\ 
Smart Grid Stability & 142.6s & 121.6s & 89.3s & 270.7s \\ 
Oil & 638.2s & 82.9s & 9.0s & 211.2s \\ 
COIL-2000 & 2910.3s & 1030.3s & 59.0s & 1867.2s \\ 
Spambase & 1338.4s & 1159.9s & 51.2s & 1056.3s \\ 
Malware Detection & 6894.7s & 4175.2s & 375.9s & 8274.9s \\ 
Performance Prediction & 138.9s & 30.3s & 19.7s & 94.9s \\ 
Churn Modelling & 215.0s & 224.3s & 186.1s & 451.2s \\ 
Credit Card Fraud & 1796.0s & 147.9s & 21.9s & 657.4s \\ 
\bottomrule 
\end{tabular} 
}
\end{table}
In stark contrast, PCAForest's performance is very consistent across datasets. For example, PCAForest needed only 82.9 seconds compared with 638.2 seconds for Forest-Diffusion for the oil dataset. This efficiency is derived from the fact that, due to the use of Principal Component Analysis (PCA), the latent space is built in a way that avoids the reconstruction overhead of an autoencoder. As all computations are carried out in lower-dimensional space, PCAForest is capable of significant time savings in processing time and thus a suitable candidate for applications on resource-limited systems.

In general, EmbedForest produces the lowest execution times among all models, being the most efficient one on the largest datasets such as Spambase, Oil, COIL-2000, Malware Detection \& Credit Card Fraud. It measured almost instantaneous augmentation with respect to competing methods in the Oil dataset (9.0s). This benefit comes from its lightweight autoencoder structure, which gives a straightforward and computationally cheap mapping from input data to latent space. By pairing this minimized autoencoder with Forest-Diffusion, EmbedForest achieves a balance between representation ability and speed, thus becoming the most computationally efficient model to date in practice.

Results show a clear model runtime-complexity trade-off between baseline Forest-Diffusion and the more complex closure models. On average, AttentionForest is 28\% slower than base Forest Diffusion, as, being transformer-based, it brings significant overhead, while PCAForest and EmbedForest provide significant gains in computational efficiency (40\% and 53\%, respectively). Practitioners should therefore weigh the tradeoff between the representational richness of transformer-based approaches and the utility benefits of simpler architectures, particularly if time or compute is limited.

\section{Conclusion}

We presented a latent-space, tree-driven diffusion framework for minority-class augmentation in tabular learning, instantiated in three variants that span fidelity–efficiency trade-offs: PCAForest, EmbedForest, and AttentionForest. The design learns the diffusion vector field with gradient-boosted trees under a flow-matching objective and performs both training and sampling in compact latent spaces, aligning the generator with inductive biases that dominate tabular prediction while keeping the pipeline practical. Across a diverse suite of binary, imbalanced datasets, we evaluated augmentation ratios from 25\% to 300\% using Random Forest and XGBoost as downstream learners and three classes of criteria: (i) utility (minority recall, precision, F1, calibration), (ii) distributional similarity (Wasserstein distance), and (iii) empirical privacy (nearest-neighbor distance ratio and distance-to-closest-record). On average, latent-space diffusion improved minority-class recall and F1 over established alternatives while maintaining competitive calibration. Gains were most consistent for moderate-to-high augmentation levels, with improvements tending to saturate at the highest ratios. Among the three variants, \textit{AttentionForest} delivered the strongest average minority-class performance and exhibited stable behavior as augmentation increased; \textit{PCAForest} and \textit{EmbedForest} achieved near-parity utility in many settings with markedly lower generation time of up to 95\%, offering favorable accuracy–efficiency trade-offs in resource-constrained scenarios.

The similarity–privacy analysis revealed method-specific profiles that practitioners can match to deployment constraints. Training in latent space generally reduced distributional gaps to real data relative to neural baselines for tabular synthesis, while empirical privacy remained comparable to tree-diffusion references for the transformer-based variant. The autoencoder-based variant favored privacy (larger nearest-neighbor and closest-record distances) at the cost of looser distributional match, whereas the PCA-based variant produced very tight distributional alignment with minimal runtime overhead (see Table \ref{tab:privacy}). These complementary behaviors suggest a tunable frontier between privacy stringency, realism of synthetic samples, and augmentation effectiveness for downstream tasks.

Ablation studies clarified robustness and sensitive knobs. Smaller embedding dimensions modestly increased average recall, indicating that compact latent manifolds can improve generalization in low-data minority regimes. Conversely, aggressive learning rates consistently degraded performance, underscoring the need for conservative optimization in flow learning. Other components showed dataset dependence: increasing diffusion steps benefited some stability-sensitive datasets; additional attention heads improved performance where long-range feature interactions were salient(see Figure \ref{img:individual_ablation}*). Overall, the methods remained robust across practical augmentation ranges and encoder capacities, but optimal settings were task-specific.

The proposed augmentation pipeline is directly applicable in engineering domains where minority events drive risk and cost: defect and anomaly detection in manufacturing quality control; early-warning systems and fault detection in energy and water infrastructure (e.g., grid stability, pump and chiller monitoring); predictive maintenance for industrial equipment; intrusion and fraud detection in cyber-physical and financial systems; and clinical risk stratification in healthcare operations analytics. In these settings, latent-space diffusion can (i) rebalance training data to improve sensitivity to rare events, (ii) generate shareable, high-utility synthetic datasets when real data access is restricted, and (iii) support rapid what-if analysis under changing class priors.

Despite these strengths, our study has several limitations: the tree-based vector field is trained without minibatching, which increases memory footprint and may limit scalability on very large tables; the transformer autoencoder scales quadratically with feature count, making very high-dimensional inputs computationally expensive; and privacy was assessed with empirical proxies (NNDR/DCR) rather than formal guarantees, alongside a focus on binary imbalance with one-hot categorical encoding. Looking ahead, promising directions include incorporating formal privacy guarantees—for example, differentially private boosting or histogram-based DP GBDT—to quantify the utility–privacy trade-off under explicit \((\varepsilon,\delta)\) budgets; extending the framework to mixed-type and multiclass tabular synthesis using learned categorical embeddings and mixed-type decoders so augmentation respects richer label structures; and coupling augmentation with imputation via validation-guided policies that jointly select augmentation ratios and imputation strength in settings where minority classes and missingness co-occur.

In sum, learning diffusion in compact latent spaces with tree-based vector fields offers a practical, tunable route to high-fidelity, privacy-aware tabular augmentation that advances minority-class prediction while accommodating real-world constraints on runtime, data access, and risk.

\clearpage

\printbibliography

@article{ghosh2024class,
  title={The class imbalance problem in deep learning},
  author={Ghosh, Kushankur and Bellinger, Colin and Corizzo, Roberto and Branco, Paula and Krawczyk, Bartosz and Japkowicz, Nathalie},
  journal={Machine Learning},
  volume={113},
  number={7},
  pages={4845--4901},
  year={2024},
  publisher={Springer}
}

@inproceedings{machado2022benchmarking,
  title={Benchmarking data augmentation techniques for tabular data},
  author={Machado, Pedro and Fernandes, Bruno and Novais, Paulo},
  booktitle={International Conference on Intelligent Data Engineering and Automated Learning},
  pages={104--112},
  year={2022},
  organization={Springer}
}

@article{perera2021one,
  title={One-class classification: A survey},
  author={Perera, Pramuditha and Oza, Poojan and Patel, Vishal M},
  journal={arXiv preprint arXiv:2101.03064},
  year={2021}
}

@article{sauber2022use,
  title={The use of generative adversarial networks to alleviate class imbalance in tabular data: a survey},
  author={Sauber-Cole, Rick and Khoshgoftaar, Taghi M},
  journal={Journal of Big Data},
  volume={9},
  number={1},
  pages={98},
  year={2022},
  publisher={Springer}
}

@article{ma2025class,
  title={Class-imbalanced learning on graphs: A survey},
  author={Ma, Yihong and Tian, Yijun and Moniz, Nuno and Chawla, Nitesh V},
  journal={ACM Computing Surveys},
  volume={57},
  number={8},
  pages={1--16},
  year={2025},
  publisher={ACM New York, NY}
}

@article{sauglam2022novel,
  title={A novel SMOTE-based resampling technique trough noise detection and the boosting procedure},
  author={Sa{\u{g}}lam, Fatih and Cengiz, Mehmet Ali},
  journal={Expert Systems with Applications},
  volume={200},
  pages={117023},
  year={2022},
  publisher={Elsevier}
}

@article{abedin2023combining,
  title={Combining weighted SMOTE with ensemble learning for the class-imbalanced prediction of small business credit risk},
  author={Abedin, Mohammad Zoynul and Guotai, Chi and Hajek, Petr and Zhang, Tong},
  journal={Complex \& Intelligent Systems},
  volume={9},
  number={4},
  pages={3559--3579},
  year={2023},
  publisher={Springer}
}

@article{dixit2023sampling,
  title={Sampling technique for noisy and borderline examples problem in imbalanced classification},
  author={Dixit, Abhishek and Mani, Ashish},
  journal={Applied Soft Computing},
  volume={142},
  pages={110361},
  year={2023},
  publisher={Elsevier}
}

@article{chen2024survey,
  title={A survey on imbalanced learning: latest research, applications and future directions},
  author={Chen, Wuxing and Yang, Kaixiang and Yu, Zhiwen and Shi, Yifan and Chen, CL Philip},
  journal={Artificial Intelligence Review},
  volume={57},
  number={6},
  pages={137},
  year={2024},
  publisher={Springer}
}

@article{tsai2024hybrid,
  title={A Hybrid approach for binary classification of imbalanced data},
  author={Tsai, Hsinhan and Yang, Ta-Wei and Wong, Wai-Man and Kao, Han-Yi and Chou, Cheng-Fu},
  journal={International Journal of Computational Intelligence and Applications},
  volume={23},
  number={03},
  pages={2450013},
  year={2024},
  publisher={World Scientific}
}

@inproceedings{bellinger2017sampling,
  title={Sampling a longer life: Binary versus one-class classification revisited},
  author={Bellinger, Colin and Sharma, Shiven and Za{\i}ane, Osmar R and Japkowicz, Nathalie},
  booktitle={First International Workshop on Learning with Imbalanced Domains: Theory and Applications},
  pages={64--78},
  year={2017},
  organization={PMLR}
}

@article{seliya2021literature,
  title={A literature review on one-class classification and its potential applications in big data},
  author={Seliya, Naeem and Abdollah Zadeh, Azadeh and Khoshgoftaar, Taghi M},
  journal={Journal of Big Data},
  volume={8},
  pages={1--31},
  year={2021},
  publisher={Springer}
}

@inproceedings{japkowicz2000learning,
  title={Learning from imbalanced data sets: a comparison of various strategies},
  author={Japkowicz, Nathalie and others},
  booktitle={AAAI workshop on learning from imbalanced data sets},
  volume={68},
  pages={10--15},
  year={2000},
  organization={AAAI Press Menlo Park, CA}
}

@inproceedings{japkowicz2003class,
  title={Class imbalances: are we focusing on the right issue},
  author={Japkowicz, Nathalie},
  booktitle={Workshop on learning from imbalanced data sets II},
  volume={1723},
  pages={63},
  year={2003}
}

@inproceedings{japkowicz1995novelty,
  title={A novelty detection approach to classification},
  author={Japkowicz, Nathalie and Myers, Catherine and Gluck, Mark and others},
  booktitle={IJCAI},
  volume={1},
  pages={518--523},
  year={1995},
  organization={Citeseer}
}

@article{chaipanha2022smote,
  title={SMOTE VS. RANDOM UNDERSAMPLING FOR IMBALANCED DATA-CAR OWNERSHIP DEMAND MODEL.},
  author={Chaipanha, Wuttikrai and Kaewwichian, Patiphan},
  journal={Komunik{\'a}cie},
  volume={24},
  number={3},
  year={2022}
}

@article{tarawneh2022stop,
  title={Stop oversampling for class imbalance learning: A review},
  author={Tarawneh, Ahmad S and Hassanat, Ahmad B and Altarawneh, Ghada Awad and Almuhaimeed, Abdullah},
  journal={IEEe Access},
  volume={10},
  pages={47643--47660},
  year={2022},
  publisher={IEEE}
}

@inproceedings{bunkhumpornpat2009safe,
  title={Safe-level-smote: Safe-level-synthetic minority over-sampling technique for handling the class imbalanced problem},
  author={Bunkhumpornpat, Chumphol and Sinapiromsaran, Krung and Lursinsap, Chidchanok},
  booktitle={Advances in knowledge discovery and data mining: 13th Pacific-Asia conference, PAKDD 2009 Bangkok, Thailand, April 27-30, 2009 proceedings 13},
  pages={475--482},
  year={2009},
  organization={Springer}
}

@article{xu2019modeling,
  title={Modeling tabular data using conditional gan},
  author={Xu, Lei and Skoularidou, Maria and Cuesta-Infante, Alfredo and Veeramachaneni, Kalyan},
  journal={Advances in neural information processing systems},
  volume={32},
  year={2019}
}

@article{engelmann2020conditional,
  title={Conditional Wasserstein GAN-based oversampling of tabular data for imbalanced learning},
  author={Engelmann, Justin and Lessmann, Stefan},
  journal={arXiv preprint arXiv:2008.09202},
  year={2020}
}

@inproceedings{zhao2021ctab,
  title={Ctab-gan: Effective table data synthesizing},
  author={Zhao, Zilong and Kunar, Aditya and Birke, Robert and Chen, Lydia Y},
  booktitle={Asian Conference on Machine Learning},
  pages={97--112},
  year={2021},
  organization={PMLR}
}

@inproceedings{chen2020gan,
  title={Gan-leaks: A taxonomy of membership inference attacks against generative models},
  author={Chen, Dingfan and Yu, Ning and Zhang, Yang and Fritz, Mario},
  booktitle={Proceedings of the 2020 ACM SIGSAC conference on computer and communications security},
  pages={343--362},
  year={2020}
}

@article{bughin2018notes,
  title={Notes from the AI frontier: Modeling the impact of AI on the world economy},
  author={Bughin, Jacques and Seong, Jeongmin and Manyika, James and Chui, Michael and Joshi, Raoul},
  journal={McKinsey Global Institute},
  volume={4},
  number={1},
  year={2018}
}

@article{nock2022generative,
  title={Generative trees: Adversarial and copycat},
  author={Nock, Richard and Guillame-Bert, Mathieu},
  journal={arXiv preprint arXiv:2201.11205},
  year={2022}
}

@article{mirza2014conditional,
  title={Conditional generative adversarial nets},
  author={Mirza, Mehdi and Osindero, Simon},
  journal={arXiv preprint arXiv:1411.1784},
  year={2014}
}

@article{fiore2019using,
  title={Using generative adversarial networks for improving classification effectiveness in credit card fraud detection},
  author={Fiore, Ugo and De Santis, Alfredo and Perla, Francesca and Zanetti, Paolo and Palmieri, Francesco},
  journal={Information Sciences},
  volume={479},
  pages={448--455},
  year={2019},
  publisher={Elsevier}
}

@inproceedings{dal2015calibrating,
  title={Calibrating probability with undersampling for unbalanced classification},
  author={Dal Pozzolo, Andrea and Caelen, Olivier and Johnson, Reid A and Bontempi, Gianluca},
  booktitle={2015 IEEE symposium series on computational intelligence},
  pages={159--166},
  year={2015},
  organization={IEEE}
}

@article{hagestedt2019mbeacon,
  title={MBeacon: Privacy-preserving beacons for DNA methylation data},
  author={Hagestedt, Inken and Zhang, Yang and Humbert, Mathias and Berrang, Pascal and Haixu, Tang and XiaoFeng, Wang and Backes, Michael},
  year={2019},
  publisher={CISPA}
}

@article{leevy2023comparative,
  title={Comparative analysis of binary and one-class classification techniques for credit card fraud data},
  author={Leevy, Joffrey L and Hancock, John and Khoshgoftaar, Taghi M},
  journal={Journal of Big Data},
  volume={10},
  number={1},
  pages={118},
  year={2023},
  publisher={Springer}
}

@inproceedings{bellinger2012one,
  title={One-class versus binary classification: Which and when?},
  author={Bellinger, Colin and Sharma, Shiven and Japkowicz, Nathalie},
  booktitle={2012 11th international conference on machine learning and applications},
  volume={2},
  pages={102--106},
  year={2012},
  organization={IEEE}
}

@inproceedings{mohammed2020machine,
  title={Machine learning with oversampling and undersampling techniques: overview study and experimental results},
  author={Mohammed, Roweida and Rawashdeh, Jumanah and Abdullah, Malak},
  booktitle={2020 11th international conference on information and communication systems (ICICS)},
  pages={243--248},
  year={2020},
  organization={IEEE}
}

@article{wongvorachan2023comparison,
  title={A comparison of undersampling, oversampling, and SMOTE methods for dealing with imbalanced classification in educational data mining},
  author={Wongvorachan, Tarid and He, Surina and Bulut, Okan},
  journal={Information},
  volume={14},
  number={1},
  pages={54},
  year={2023},
  publisher={MDPI}
}

@article{mease2007boosted,
  title={Boosted classification trees and class probability/quantile estimation.},
  author={Mease, David and Wyner, Abraham J and Buja, Andreas},
  journal={Journal of Machine Learning Research},
  volume={8},
  number={3},
  year={2007}
}

@article{ganganwar2012overview,
  title={An overview of classification algorithms for imbalanced datasets},
  author={Ganganwar, Vaishali},
  journal={International Journal of Emerging Technology and Advanced Engineering},
  volume={2},
  number={4},
  pages={42--47},
  year={2012}
}

@inproceedings{drummond2003c4,
  title={C4. 5, class imbalance, and cost sensitivity: why under-sampling beats over-sampling},
  author={Drummond, Chris and Holte, Robert C and others},
  booktitle={Workshop on learning from imbalanced datasets II},
  volume={11},
  number={1--8},
  year={2003}
}

@article{santos2018cross,
  title={Cross-validation for imbalanced datasets: avoiding overoptimistic and overfitting approaches [research frontier]},
  author={Santos, Miriam Seoane and Soares, Jastin Pompeu and Abreu, Pedro Henrigues and Araujo, Helder and Santos, Joao},
  journal={ieee ComputatioNal iNtelligeNCe magaziNe},
  volume={13},
  number={4},
  pages={59--76},
  year={2018},
  publisher={IEEE}
}

@inproceedings{shamsudin2020combining,
  title={Combining oversampling and undersampling techniques for imbalanced classification: A comparative study using credit card fraudulent transaction dataset},
  author={Shamsudin, Haziqah and Yusof, Umi Kalsom and Jayalakshmi, Andal and Khalid, Mohd Nor Akmal},
  booktitle={2020 IEEE 16th international conference on control \& automation (ICCA)},
  pages={803--808},
  year={2020},
  organization={IEEE}
}

@article{loyola2016study,
  title={Study of the impact of resampling methods for contrast pattern based classifiers in imbalanced databases},
  author={Loyola-Gonz{\'a}lez, Octavio and Mart{\'\i}nez-Trinidad, Jos{\'e} Fco and Carrasco-Ochoa, Jes{\'u}s Ariel and Garc{\'\i}a-Borroto, Milton},
  journal={Neurocomputing},
  volume={175},
  pages={935--947},
  year={2016},
  publisher={Elsevier}
}

@article{chawla2002smote,
  title={SMOTE: synthetic minority over-sampling technique},
  author={Chawla, Nitesh V and Bowyer, Kevin W and Hall, Lawrence O and Kegelmeyer, W Philip},
  journal={Journal of artificial intelligence research},
  volume={16},
  pages={321--357},
  year={2002}
}

@article{fernandez2018smote,
  title={SMOTE for learning from imbalanced data: progress and challenges, marking the 15-year anniversary},
  author={Fern{\'a}ndez, Alberto and Garcia, Salvador and Herrera, Francisco and Chawla, Nitesh V},
  journal={Journal of artificial intelligence research},
  volume={61},
  pages={863--905},
  year={2018}
}

@article{branco2016survey,
  title={A survey of predictive modeling on imbalanced domains},
  author={Branco, Paula and Torgo, Lu{\'\i}s and Ribeiro, Rita P},
  journal={ACM computing surveys (CSUR)},
  volume={49},
  number={2},
  pages={1--50},
  year={2016},
  publisher={ACM New York, NY, USA}
}

@inproceedings{han2005borderline,
  title={Borderline-SMOTE: a new over-sampling method in imbalanced data sets learning},
  author={Han, Hui and Wang, Wen-Yuan and Mao, Bing-Huan},
  booktitle={International conference on intelligent computing},
  pages={878--887},
  year={2005},
  organization={Springer}
}

@inproceedings{he2008adasyn,
  title={ADASYN: Adaptive synthetic sampling approach for imbalanced learning},
  author={He, Haibo and Bai, Yang and Garcia, Edwardo A and Li, Shutao},
  booktitle={2008 IEEE international joint conference on neural networks (IEEE world congress on computational intelligence)},
  pages={1322--1328},
  year={2008},
  organization={Ieee}
}

@article{goodfellow2014generative,
  title={Generative adversarial nets},
  author={Goodfellow, Ian J and Pouget-Abadie, Jean and Mirza, Mehdi and Xu, Bing and Warde-Farley, David and Ozair, Sherjil and Courville, Aaron and Bengio, Yoshua},
  journal={Advances in neural information processing systems},
  volume={27},
  year={2014}
}

@article{lucini2022real,
  title={The real deal about synthetic data},
  author={Lucini, Fernando},
  journal={MIT Sloan Management Review},
  volume={63},
  number={2},
  pages={11--13},
  year={2022},
  publisher={Massachusetts Institute of Technology, Cambridge, MA}
}

@inproceedings{mescheder2018training,
  title={Which training methods for GANs do actually converge?},
  author={Mescheder, Lars and Geiger, Andreas and Nowozin, Sebastian},
  booktitle={International conference on machine learning},
  pages={3481--3490},
  year={2018},
  organization={PMLR}
}

@inproceedings{zhang2019towards,
  title={Towards the gradient vanishing, divergence mismatching and mode collapse of generative adversarial nets},
  author={Zhang, Zhaoyu and Luo, Changwei and Yu, Jun},
  booktitle={Proceedings of the 28th ACM International Conference on Information and Knowledge Management},
  pages={2377--2380},
  year={2019}
}

@article{dhariwal2021diffusion,
  title={Diffusion models beat gans on image synthesis},
  author={Dhariwal, Prafulla and Nichol, Alexander},
  journal={Advances in neural information processing systems},
  volume={34},
  pages={8780--8794},
  year={2021}
}

@article{sharma2024generative,
  title={Generative adversarial networks (GANs): introduction, taxonomy, variants, limitations, and applications},
  author={Sharma, Preeti and Kumar, Manoj and Sharma, Hitesh Kumar and Biju, Soly Mathew},
  journal={Multimedia Tools and Applications},
  pages={1--48},
  year={2024},
  publisher={Springer}
}

@article{child2020very,
  title={Very deep vaes generalize autoregressive models and can outperform them on images},
  author={Child, Rewon},
  journal={arXiv preprint arXiv:2011.10650},
  year={2020}
}

@article{de2019hierarchical,
  title={Hierarchical autoregressive image models with auxiliary decoders},
  author={De Fauw, Jeffrey and Dieleman, Sander and Simonyan, Karen},
  journal={arXiv preprint arXiv:1903.04933},
  year={2019}
}

@article{nash2021generating,
  title={Generating images with sparse representations},
  author={Nash, Charlie and Menick, Jacob and Dieleman, Sander and Battaglia, Peter W},
  journal={arXiv preprint arXiv:2103.03841},
  year={2021}
}

@inproceedings{nichol2021improved,
  title={Improved denoising diffusion probabilistic models},
  author={Nichol, Alexander Quinn and Dhariwal, Prafulla},
  booktitle={International conference on machine learning},
  pages={8162--8171},
  year={2021},
  organization={PMLR}
}

@article{franzese2023much,
  title={How much is enough? a study on diffusion times in score-based generative models},
  author={Franzese, Giulio and Rossi, Simone and Yang, Lixuan and Finamore, Alessandro and Rossi, Dario and Filippone, Maurizio and Michiardi, Pietro},
  journal={Entropy},
  volume={25},
  number={4},
  pages={633},
  year={2023},
  publisher={MDPI}
}

@article{yang2023diffusion,
  title={Diffusion models: A comprehensive survey of methods and applications},
  author={Yang, Ling and Zhang, Zhilong and Song, Yang and Hong, Shenda and Xu, Runsheng and Zhao, Yue and Zhang, Wentao and Cui, Bin and Yang, Ming-Hsuan},
  journal={ACM Computing Surveys},
  volume={56},
  number={4},
  pages={1--39},
  year={2023},
  publisher={ACM New York, NY, USA}
}

@article{cao2024survey,
  title={A survey on generative diffusion models},
  author={Cao, Hanqun and Tan, Cheng and Gao, Zhangyang and Xu, Yilun and Chen, Guangyong and Heng, Pheng-Ann and Li, Stan Z},
  journal={IEEE Transactions on Knowledge and Data Engineering},
  year={2024},
  publisher={IEEE}
}

@article{kadra2021well,
  title={Well-tuned simple nets excel on tabular datasets},
  author={Kadra, Arlind and Lindauer, Marius and Hutter, Frank and Grabocka, Josif},
  journal={Advances in neural information processing systems},
  volume={34},
  pages={23928--23941},
  year={2021}
}

@article{jdanov2019human,
  title={Human mortality database},
  author={Jdanov, Dmitri A and Jasilionis, Domantas and Shkolnikov, Vladimir M and Barbieri, Magali},
  journal={Encyclopedia of gerontology and population aging},
  pages={1--8},
  year={2019},
  publisher={Springer International Publishing}
}

@article{borisov2022deep,
  title={Deep neural networks and tabular data: A survey},
  author={Borisov, Vadim and Leemann, Tobias and Se{\ss}ler, Kathrin and Haug, Johannes and Pawelczyk, Martin and Kasneci, Gjergji},
  journal={IEEE transactions on neural networks and learning systems},
  year={2022},
  publisher={IEEE}
}

@book{ryan2020deep,
  title={Deep learning with structured data},
  author={Ryan, Mark},
  year={2020},
  publisher={Manning}
}

@article{pang2021deep,
  title={Deep learning for anomaly detection: A review},
  author={Pang, Guansong and Shen, Chunhua and Cao, Longbing and Hengel, Anton Van Den},
  journal={ACM computing surveys (CSUR)},
  volume={54},
  number={2},
  pages={1--38},
  year={2021},
  publisher={ACM New York, NY, USA}
}

@article{sanchez2020improving,
  title={Improving deep learning performance with missing values via deletion and compensation},
  author={S{\'a}nchez-Morales, Adri{\'a}n and Sancho-G{\'o}mez, Jos{\'e}-Luis and Mart{\'\i}nez-Garc{\'\i}a, Juan-Antonio and Figueiras-Vidal, An{\'\i}bal R},
  journal={Neural Computing and Applications},
  volume={32},
  pages={13233--13244},
  year={2020},
  publisher={Springer}
}

@article{borisov2023deeptlf,
  title={DeepTLF: robust deep neural networks for heterogeneous tabular data},
  author={Borisov, Vadim and Broelemann, Klaus and Kasneci, Enkelejda and Kasneci, Gjergji},
  journal={International Journal of Data Science and Analytics},
  volume={16},
  number={1},
  pages={85--100},
  year={2023},
  publisher={Springer}
}

@article{hancock2020survey,
  title={Survey on categorical data for neural networks},
  author={Hancock, John T and Khoshgoftaar, Taghi M},
  journal={Journal of big data},
  volume={7},
  number={1},
  pages={28},
  year={2020},
  publisher={Springer}
}

@article{fakoor2020fast,
  title={Fast, accurate, and simple models for tabular data via augmented distillation},
  author={Fakoor, Rasool and Mueller, Jonas W and Erickson, Nick and Chaudhari, Pratik and Smola, Alexander J},
  journal={Advances in Neural Information Processing Systems},
  volume={33},
  pages={8671--8681},
  year={2020}
}

@inproceedings{sattarov2023findiff,
  title={Findiff: Diffusion models for financial tabular data generation},
  author={Sattarov, Timur and Schreyer, Marco and Borth, Damian},
  booktitle={Proceedings of the Fourth ACM International Conference on AI in Finance},
  pages={64--72},
  year={2023}
}

@article{ho2020denoising,
  title={Denoising diffusion probabilistic models},
  author={Ho, Jonathan and Jain, Ajay and Abbeel, Pieter},
  journal={Advances in neural information processing systems},
  volume={33},
  pages={6840--6851},
  year={2020}
}

@inproceedings{sohl2015deep,
  title={Deep unsupervised learning using nonequilibrium thermodynamics},
  author={Sohl-Dickstein, Jascha and Weiss, Eric and Maheswaranathan, Niru and Ganguli, Surya},
  booktitle={International conference on machine learning},
  pages={2256--2265},
  year={2015},
  organization={pmlr}
}

@article{song2020score,
  title={Score-based generative modeling through stochastic differential equations},
  author={Song, Yang and Sohl-Dickstein, Jascha and Kingma, Diederik P and Kumar, Abhishek and Ermon, Stefano and Poole, Ben},
  journal={arXiv preprint arXiv:2011.13456},
  year={2020}
}

@article{baranchuk2021label,
  title={Label-efficient semantic segmentation with diffusion models},
  author={Baranchuk, Dmitry and Rubachev, Ivan and Voynov, Andrey and Khrulkov, Valentin and Babenko, Artem},
  journal={arXiv preprint arXiv:2112.03126},
  year={2021}
}

@inproceedings{kotelnikov2023tabddpm,
  title={Tabddpm: Modelling tabular data with diffusion models},
  author={Kotelnikov, Akim and Baranchuk, Dmitry and Rubachev, Ivan and Babenko, Artem},
  booktitle={International Conference on Machine Learning},
  pages={17564--17579},
  year={2023},
  organization={PMLR}
}

@inproceedings{jolicoeur2024generating,
  title={Generating and imputing tabular data via diffusion and flow-based gradient-boosted trees},
  author={Jolicoeur-Martineau, Alexia and Fatras, Kilian and Kachman, Tal},
  booktitle={International Conference on Artificial Intelligence and Statistics},
  pages={1288--1296},
  year={2024},
  organization={PMLR}
}

@article{shwartz2022tabular,
  title={Tabular data: Deep learning is not all you need},
  author={Shwartz-Ziv, Ravid and Armon, Amitai},
  journal={Information Fusion},
  volume={81},
  pages={84--90},
  year={2022},
  publisher={Elsevier}
}

@article{saharia2022photorealistic,
  title={Photorealistic text-to-image diffusion models with deep language understanding},
  author={Saharia, Chitwan and Chan, William and Saxena, Saurabh and Li, Lala and Whang, Jay and Denton, Emily L and Ghasemipour, Kamyar and Gontijo Lopes, Raphael and Karagol Ayan, Burcu and Salimans, Tim and others},
  journal={Advances in neural information processing systems},
  volume={35},
  pages={36479--36494},
  year={2022}
}

@article{abdi2010principal,
  title={Principal component analysis},
  author={Abdi, Herv{\'e} and Williams, Lynne J},
  journal={Wiley interdisciplinary reviews: computational statistics},
  volume={2},
  number={4},
  pages={433--459},
  year={2010},
  publisher={Wiley Online Library}
}

@inproceedings{rombach2022high,
  title={High-resolution image synthesis with latent diffusion models},
  author={Rombach, Robin and Blattmann, Andreas and Lorenz, Dominik and Esser, Patrick and Ommer, Bj{\"o}rn},
  booktitle={Proceedings of the IEEE/CVF conference on computer vision and pattern recognition},
  pages={10684--10695},
  year={2022}
}

@article{lovelace2023latent,
  title={Latent diffusion for language generation},
  author={Lovelace, Justin and Kishore, Varsha and Wan, Chao and Shekhtman, Eliot and Weinberger, Kilian Q},
  journal={Advances in Neural Information Processing Systems},
  volume={36},
  pages={56998--57025},
  year={2023}
}

@article{liu2023audioldm,
  title={Audioldm: Text-to-audio generation with latent diffusion models},
  author={Liu, Haohe and Chen, Zehua and Yuan, Yi and Mei, Xinhao and Liu, Xubo and Mandic, Danilo and Wang, Wenwu and Plumbley, Mark D},
  journal={arXiv preprint arXiv:2301.12503},
  year={2023}
}

@article{asperti2023comparing,
  title={Comparing the latent space of generative models},
  author={Asperti, Andrea and Tonelli, Valerio},
  journal={Neural Computing and Applications},
  volume={35},
  number={4},
  pages={3155--3172},
  year={2023},
  publisher={Springer}
}

@article{mittal2021symbolic,
  title={Symbolic music generation with diffusion models},
  author={Mittal, Gautam and Engel, Jesse and Hawthorne, Curtis and Simon, Ian},
  journal={arXiv preprint arXiv:2103.16091},
  year={2021}
}

@article{vahdat2021score,
  title={Score-based generative modeling in latent space. 2021},
  author={Vahdat, Arash and Kreis, Karsten and Kautz, Jan},
  journal={URL https://arxiv. org/abs/2106.05931},
  year={2021}
}

@misc{kingma2013auto,
  title={Auto-encoding variational bayes},
  author={Kingma, Diederik P and Welling, Max and others},
  year={2013},
  publisher={Banff, Canada}
}

@article{yoon2018personalized,
  title={Personalized survival predictions via Trees of Predictors: An application to cardiac transplantation},
  author={Yoon, Jinsung and Zame, William R and Banerjee, Amitava and Cadeiras, Martin and Alaa, Ahmed M and van der Schaar, Mihaela},
  journal={PloS one},
  volume={13},
  number={3},
  pages={e0194985},
  year={2018},
  publisher={Public Library of Science San Francisco, CA USA}
}

@article{emmanuel2021survey,
  title={A survey on missing data in machine learning},
  author={Emmanuel, Tlamelo and Maupong, Thabiso and Mpoeleng, Dimane and Semong, Thabo and Mphago, Banyatsang and Tabona, Oteng},
  journal={Journal of Big data},
  volume={8},
  pages={1--37},
  year={2021},
  publisher={Springer}
}

@article{ouyang2023missdiff,
  title={Missdiff: Training diffusion models on tabular data with missing values},
  author={Ouyang, Yidong and Xie, Liyan and Li, Chongxuan and Cheng, Guang},
  journal={arXiv preprint arXiv:2307.00467},
  year={2023}
}

@article{bertsimas2021simple,
  title={Simple Imputation Rules for Prediction with Missing Data: Contrasting Theoretical Guarantees with Empirical Performance},
  author={Bertsimas, Dimitris and Delarue, Arthur and Pauphilet, Jean},
  journal={arXiv preprint arXiv:2104.03158},
  year={2021}
}

@article{jolicoeur2021gotta,
  title={Gotta go fast when generating data with score-based models},
  author={Jolicoeur-Martineau, Alexia and Li, Ke and Pich{\'e}-Taillefer, R{\'e}mi and Kachman, Tal and Mitliagkas, Ioannis},
  journal={arXiv preprint arXiv:2105.14080},
  year={2021}
}

@article{san2021noise,
  title={Noise estimation for generative diffusion models},
  author={San-Roman, Robin and Nachmani, Eliya and Wolf, Lior},
  journal={arXiv preprint arXiv:2104.02600},
  year={2021}
}

@article{zhang2023mixed,
  title={Mixed-type tabular data synthesis with score-based diffusion in latent space},
  author={Zhang, Hengrui and Zhang, Jiani and Srinivasan, Balasubramaniam and Shen, Zhengyuan and Qin, Xiao and Faloutsos, Christos and Rangwala, Huzefa and Karypis, George},
  journal={arXiv preprint arXiv:2310.09656},
  year={2023}
}

@article{song2020denoising,
  title={Denoising diffusion implicit models},
  author={Song, Jiaming and Meng, Chenlin and Ermon, Stefano},
  journal={arXiv preprint arXiv:2010.02502},
  year={2020}
}

@book{watt2020machine,
  title={Machine learning refined: Foundations, algorithms, and applications},
  author={Watt, Jeremy and Borhani, Reza and Katsaggelos, Aggelos K},
  year={2020},
  publisher={Cambridge University Press}
}

@article{grinsztajn2022tree,
  title={Why do tree-based models still outperform deep learning on typical tabular data?},
  author={Grinsztajn, L{\'e}o and Oyallon, Edouard and Varoquaux, Ga{\"e}l},
  journal={Advances in neural information processing systems},
  volume={35},
  pages={507--520},
  year={2022}
}

@article{blockeel2023decision,
  title={Decision trees: from efficient prediction to responsible AI},
  author={Blockeel, Hendrik and Devos, Laurens and Fr{\'e}nay, Beno{\^\i}t and Nanfack, G{\'e}raldin and Nijssen, Siegfried},
  journal={Frontiers in artificial intelligence},
  volume={6},
  pages={1124553},
  year={2023},
  publisher={Frontiers Media SA}
}

@article{ma2020diagnostic,
  title={Diagnostic classification of cancers using extreme gradient boosting algorithm and multi-omics data},
  author={Ma, Baoshan and Meng, Fanyu and Yan, Ge and Yan, Haowen and Chai, Bingjie and Song, Fengju},
  journal={Computers in biology and medicine},
  volume={121},
  pages={103761},
  year={2020},
  publisher={Elsevier}
}

@article{florek2023benchmarking,
  title={Benchmarking state-of-the-art gradient boosting algorithms for classification},
  author={Florek, Piotr and Zagda{\'n}ski, Adam},
  journal={arXiv preprint arXiv:2305.17094},
  year={2023}
}

@article{tong2023improving,
  title={Improving and generalizing flow-based generative models with minibatch optimal transport},
  author={Tong, Alexander and Fatras, Kilian and Malkin, Nikolay and Huguet, Guillaume and Zhang, Yanlei and Rector-Brooks, Jarrid and Wolf, Guy and Bengio, Yoshua},
  journal={arXiv preprint arXiv:2302.00482},
  year={2023}
}

@article{vaswani2017attention,
  title={Attention is all you need},
  author={Vaswani, Ashish and Shazeer, Noam and Parmar, Niki and Uszkoreit, Jakob and Jones, Llion and Gomez, Aidan N and Kaiser, {\L}ukasz and Polosukhin, Illia},
  journal={Advances in neural information processing systems},
  volume={30},
  year={2017}
}

@article{cheng2024novel,
  title={A novel transformer autoencoder for multi-modal emotion recognition with incomplete data},
  author={Cheng, Cheng and Liu, Wenzhe and Fan, Zhaoxin and Feng, Lin and Jia, Ziyu},
  journal={Neural Networks},
  volume={172},
  pages={106111},
  year={2024},
  publisher={Elsevier}
}

@article{fonseca2023tabular,
  title={Tabular and latent space synthetic data generation: a literature review},
  author={Fonseca, Joao and Bacao, Fernando},
  journal={Journal of Big Data},
  volume={10},
  number={1},
  pages={115},
  year={2023},
  publisher={Springer}
}

@article{kim2022stasy,
  title={Stasy: Score-based tabular data synthesis},
  author={Kim, Jayoung and Lee, Chaejeong and Park, Noseong},
  journal={arXiv preprint arXiv:2210.04018},
  year={2022}
}

@inproceedings{schramowski2023safe,
  title={Safe latent diffusion: Mitigating inappropriate degeneration in diffusion models},
  author={Schramowski, Patrick and Brack, Manuel and Deiseroth, Bj{\"o}rn and Kersting, Kristian},
  booktitle={Proceedings of the IEEE/CVF Conference on Computer Vision and Pattern Recognition},
  pages={22522--22531},
  year={2023}
}

@article{yang2024balanced,
  title={Balanced mixed-type tabular data synthesis with diffusion models},
  author={Yang, Zeyu and Yu, Han and Guo, Peikun and Zanna, Khadija and Yang, Xiaoxue and Sano, Akane},
  journal={arXiv preprint arXiv:2404.08254},
  year={2024}
}

@inproceedings{lee2023codi,
  title={Codi: Co-evolving contrastive diffusion models for mixed-type tabular synthesis},
  author={Lee, Chaejeong and Kim, Jayoung and Park, Noseong},
  booktitle={International Conference on Machine Learning},
  pages={18940--18956},
  year={2023},
  organization={PMLR}
}

@article{suh2023autodiff,
  title={Autodiff: combining auto-encoder and diffusion model for tabular data synthesizing},
  author={Suh, Namjoon and Lin, Xiaofeng and Hsieh, Din-Yin and Honarkhah, Merhdad and Cheng, Guang},
  journal={arXiv preprint arXiv:2310.15479},
  year={2023}
}

@inproceedings{shang2023precision,
  title={Precision/recall on imbalanced test data},
  author={Shang, Hongwei and Langlois, Jean-Marc and Tsioutsiouliklis, Kostas and Kang, Changsung},
  booktitle={International Conference on Artificial Intelligence and Statistics},
  pages={9879--9891},
  year={2023},
  organization={PMLR}
}

@article{miao2022precision,
  title={Precision--recall curve (PRC) classification trees},
  author={Miao, Jiaju and Zhu, Wei},
  journal={Evolutionary intelligence},
  volume={15},
  number={3},
  pages={1545--1569},
  year={2022},
  publisher={Springer}
}

@article{bomze2023optimization,
  title={Optimization under uncertainty and risk: Quadratic and copositive approaches},
  author={Bomze, Immanuel M and Gabl, Markus},
  journal={European Journal of Operational Research},
  volume={310},
  number={2},
  pages={449--476},
  year={2023},
  publisher={Elsevier}
}

@article{leo2023wasserstein,
  title={Wasserstein distance in deep learning},
  author={Leo, Junior and Ge, Ernest and Li, Stotle},
  journal={Available at SSRN 4368733},
  year={2023}
}

@inproceedings{liu2024scaling,
  title={Scaling while privacy preserving: A comprehensive synthetic tabular data generation and evaluation in learning analytics},
  author={Liu, Qinyi and Khalil, Mohammad and Jovanovic, Jelena and Shakya, Ronas},
  booktitle={Proceedings of the 14th Learning Analytics and Knowledge Conference},
  pages={620--631},
  year={2024}
}

@article{franti2023soft,
  title={Soft precision and recall},
  author={Fr{\"a}nti, Pasi and Mariescu-Istodor, Radu},
  journal={Pattern Recognition Letters},
  volume={167},
  pages={115--121},
  year={2023},
  publisher={Elsevier}
}

@inproceedings{parmar2018review,
  title={A review on random forest: An ensemble classifier},
  author={Parmar, Aakash and Katariya, Rakesh and Patel, Vatsal},
  booktitle={International conference on intelligent data communication technologies and internet of things},
  pages={758--763},
  year={2018},
  organization={Springer}
}

@inproceedings{chen2016xgboost,
  title={Xgboost: A scalable tree boosting system},
  author={Chen, Tianqi and Guestrin, Carlos},
  booktitle={Proceedings of the 22nd acm sigkdd international conference on knowledge discovery and data mining},
  pages={785--794},
  year={2016}
}

@article{li2025improved,
  title={An improved SMOTE algorithm for enhanced imbalanced data classification by expanding sample generation space},
  author={Li, Ying and Yang, Yali and Song, Peihua and Duan, Lian and Ren, Rui},
  journal={Scientific Reports},
  volume={15},
  number={1},
  pages={23521},
  year={2025},
  publisher={Nature Publishing Group UK London}
}

@article{maldonado2019alternative,
  title={An alternative SMOTE oversampling strategy for high-dimensional datasets},
  author={Maldonado, Sebasti{\'a}n and L{\'o}pez, Julio and Vairetti, Carla},
  journal={Applied Soft Computing},
  volume={76},
  pages={380--389},
  year={2019},
  publisher={Elsevier}
}

@article{li2024accelerating,
  title={Accelerating convergence of score-based diffusion models, provably},
  author={Li, Gen and Huang, Yu and Efimov, Timofey and Wei, Yuting and Chi, Yuejie and Chen, Yuxin},
  journal={arXiv preprint arXiv:2403.03852},
  year={2024}
}

@article{asru2025automation,
  title={From automation to autonomy in smart manufacturing: a Bayesian optimization framework for modeling multi-objective experimentation and sequential decision making},
  author={Asru, Avijit Saha and Khosravi, Hamed and Ahmed, Imtiaz and Azeem, Abdullahil},
  journal={The International Journal of Advanced Manufacturing Technology},
  volume={137},
  number={9},
  pages={5027--5057},
  year={2025},
  publisher={Springer}
}

\end{document}